\documentclass[fleqn,10pt]{wlscirep}

\usepackage[utf8]{inputenc}
\usepackage[T1]{fontenc}

\usepackage{amsmath}
\usepackage{amsfonts}
\usepackage{amssymb}

\usepackage{graphicx}
\usepackage{booktabs}
\usepackage{multirow}

\usepackage{xcolor}
\usepackage{colortbl}
\usepackage{pifont}

\usepackage{bm}
\usepackage[ruled]{algorithm2e}

\SetAlFnt{\small}
\SetAlCapFnt{\small}
\SetAlCapNameFnt{\small}
\SetAlCapHSkip{0pt}

\usepackage{placeins}
\usepackage{dblfloatfix}

\usepackage[normalem]{ulem}

\usepackage{hyperref}
\usepackage{twemojis}
\usepackage{tikz}
\usepackage{dsfont}
\usepackage{makecell}

\usepackage{newtxtext,newtxmath}
\usepackage[font=footnotesize]{caption}

\usepackage{cleveref}

\usepackage[percent]{overpic}
\usepackage{siunitx,tabularx}

\AtBeginDocument{%
  \crefname{figure}{Fig.}{Figs.}%
  \Crefname{figure}{Figure}{Figures}%
  \crefname{table}{Table}{Tables}%
  \Crefname{table}{Table}{Tables}%
  \crefname{section}{Sec.}{Secs.}%
  \Crefname{section}{Section}{Sections}%
  \crefname{appendix}{Appendix}{Appendices}%
  \Crefname{appendix}{Appendix}{Appendices}%
  \crefname{equation}{Eq.}{Eqs.}%
  \Crefname{equation}{Equation}{Equations}%
}

\newif\ifsubmissioncopy
\submissioncopyfalse  % working version with colored comments

\definecolor{myorange}{RGB}{255, 165, 0}
\definecolor{mydarkorange}{RGB}{230, 165, 0}
\definecolor{mybrown}{RGB}{155, 78, 20}
\definecolor{mypurple}{RGB}{150, 56, 226}
\definecolor{myblue}{RGB}{10, 50, 230}
\definecolor{myred}{RGB}{230, 10, 10}
\definecolor{mygreen}{RGB}{60, 160, 60}
\definecolor{mydarkgreen}{RGB}{30, 110, 30}
\definecolor{myskyblue}{RGB}{31, 119, 180}
\definecolor{mymagenta}{RGB}{139, 0, 139}
\definecolor{mycyan}{RGB}{0, 100, 100}
\definecolor{winered}{RGB}{178, 18, 38}
\definecolor{mygray}{RGB}{100, 100, 100}

\newcommand{\winner}[1]{\textcolor{black}{\boldmath#1}}
\newcommand{\runnerup}[1]{\underline{#1}}

\newcommand{\etal}{\textit{et al. }}

\ifsubmissioncopy

    \newcommand{\bhargav}[1]{}
    \newcommand{\ad}[1]{}
    \newcommand{\anirudh}[1]{}
    \newcommand{\abh}[1]{}
    \newcommand{\aloo}[1]{}
    \newcommand{\pablo}[1]{}
    \newcommand{\ashok}[1]{}
    \newcommand{\todo}[1]{}

\else

    \newcommand{\bhargav}[1]{{\color{mygreen}{[BG: #1]}}}
    \newcommand{\ad}[1]{{\color{myskyblue}{[AD: #1]}}}
    \newcommand{\anirudh}[1]{{\color{winered}{[ABH: #1]}}}
    \newcommand{\abh}[1]{{\color{winered}{[ABH: #1]}}}
    \newcommand{\aloo}[1]{{\color{teal}{[AK: #1]}}}
    \newcommand{\pablo}[1]{{\color{mypurple}{[PV: #1]}}}
    \newcommand{\ashok}[1]{{\color{mymagenta}{[AV: #1]}}}

    \newcommand{\todo}[1]{\textcolor{red}{#1}}

\fi

\newcommand{\beginsupplement}{%
    \setcounter{table}{0}
    \renewcommand{\thetable}{S\arabic{table}}%
    \setcounter{figure}{0}
    \renewcommand{\thefigure}{S\arabic{figure}}%
    \setcounter{section}{0}
    \renewcommand{\thesection}{S\arabic{section}}%
}

\usepackage{twemojis}
\usepackage{tikz}

\tikzset{laser/.pic={
  \draw[red, opacity=.25, line width=3.2pt, line cap=round] (0.72,0) -- (1.25,0);
  \draw[red!85!white, line width=1.1pt] (0.72,0) -- (1.25,0);
  \fill[red, opacity=.18] (1.25,0) circle (0.13);
  \fill[red, opacity=.35] (1.25,0) circle (0.085);
  \fill[red!90!black] (1.25,0) circle (0.05);
  \fill[white, opacity=.8] (1.235,0.015) circle (0.015);
  \shade[top color=gray!25, bottom color=gray!70, rounded corners=1.2pt]
    (0,-0.1) rectangle (0.64,0.1);
  \shade[top color=gray!55, bottom color=black!70, rounded corners=1.2pt]
    (-0.02,-0.105) rectangle (0.1,0.105);
  \fill[black!60] (0.14,-0.1) rectangle (0.17,0.1);
  \shade[top color=gray!50, bottom color=black!75]
    (0.62,-0.1) -- (0.72,-0.06) -- (0.72,0.06) -- (0.62,0.1) -- cycle;
  \fill[red!80!black, rounded corners=0.6pt] (0.33,0.09) rectangle (0.45,0.13);
  \fill[white, opacity=.5] (0.1,0.045) rectangle (0.62,0.065);
}}
\title{eSPG: The Unreasonable Effectiveness of Event Speckle Plethysmography}

\author[1,$\dagger$]{Aniket Dashpute}
\author[1,$\dagger$]{Anirudh Bindiganavale Harish}
\author[1,$\dagger$]{Bhargav Ghanekar}
\author[1,2]{Alankar Kotwal}
\author[1]{Andrew Negrut}
\author[1]{Syna Gupta}
\author[2]{Pablo A. Valdes}
\author[1,*]{Ashok Veeraraghavan}
\affil[1]{Department of Electrical and Computer Engineering, Rice University, Houston, Texas, USA}
\affil[2]{Department of Neurosurgery, University of Texas Medical Branch, Galveston, Texas, USA}

\affil[$\dagger$]{contributed equally}

\affil[*]{vashok@rice.edu}

\begin{abstract}

Non-contact measurement of vital signs such as heart rate (HR) is valuable for patient monitoring and rescue operations. However, existing methods struggle with ambient illumination and motion, and have been demonstrated only at distances of tens of meters. Here we introduce event speckle plethysmography (eSPG), which illuminates a body part with a low-power (0.5-5 mW) laser and captures the resulting dynamic speckle pattern using an event camera. Event cameras, with their high dynamic range, temporal resolution, and sensitivity to intensity changes, are well-suited to capturing rapid speckle fluctuations that encode the physiological signal, even under challenging illumination and imaging conditions. We demonstrate HR estimation at 1 km with a mean absolute error of 5 beats per minute, and 90\% of estimates satisfy an AAMI-defined HR estimation criterion.
This distance exceeds previous remote plethysmography demonstrations by an order of magnitude and previous speckle plethysmography demonstrations by over three orders of magnitude. We also study and characterize robustness to motion and ambient illumination. Our theoretical analysis and experimental results demonstrate that resolving dynamic speckle fluctuations enables accurate heart rate estimation even at kilometer-scale distances, establishing a new direction toward robust, non-contact physiological monitoring.
\end{abstract}

\begin{document}

\flushbottom
\maketitle

% \linenumbers
\thispagestyle{empty}

% this line prevents the main paper sections from appearing in table of contents
\addtocontents{toc}{\protect\setcounter{tocdepth}{-1}}

% \section{Introduction}\label{sec:intro}

\begin{figure}[!t]
    \centering 
    \includegraphics[width=\linewidth]{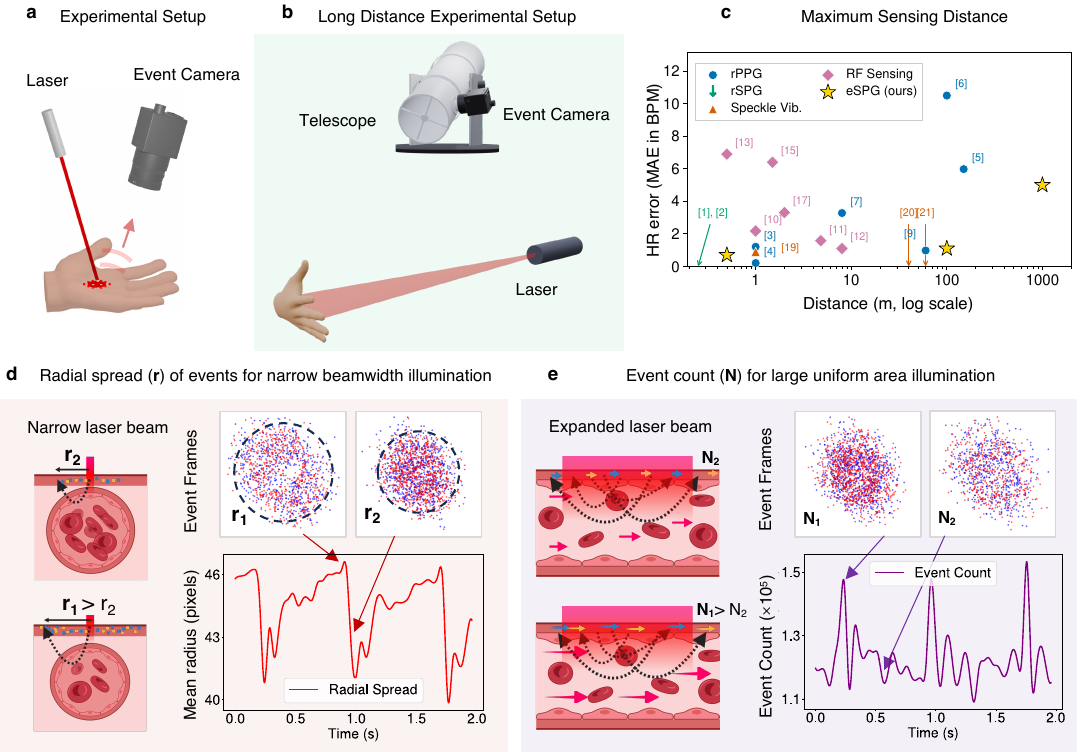}
    \caption{
    \textbf{Our approach uses laser illumination and an event camera to image speckle dynamics for plethysmography, enabling heart rate estimation at 1 km imaging distance.}
    \textbf{a,} Across all experiments, our setup comprises a laser and an event camera for capturing speckle patterns.
    \textbf{b,} Long-distance measurements use telescope optics with an event camera, while the laser illuminates from 60 to 100~m distance.
    \textbf{c,} The estimated eSPG signal is effective even at distances of $1~km$, an order of magnitude greater than state-of-the-art works across different methods (papers from the plot:  \cite{herranz2022camera, herranz2023comparing, chen2018deepphys, sun2024contrast, blackford2017measurements, blackford2016long, kong2020robust, rao2025measurement, al2017remote, vilesov2022blending, khan2022contactless, adib2015smart, zhu2024camera, wang2020csi, mura2025trade, wang2023vital, mao2024mmhrr, li2025medusa, que2022contactless, ozana2015demonstration, zalevsky2009simultaneous}; papers marked with arrows do not report heart rate estimation errors).
    \textbf{d,} When the beamwidth is narrow (e.g., in cases of short-distance illumination), a larger volume of blood from arterial dilation increases scattering, limiting the depth of light penetration, thus reducing the radial spread of the events.
    \textbf{e,} When the beamwidth is larger (e.g., in cases of long-distance illumination), the main indicator of pulsing is observed in changes in event rate, primarily due to changes in blood flow speed.
    }
    \label{fig:teaser}
\end{figure}

\noindent
Heart rate (pulse rate) is a fundamental physiological signal commonly measured through changes in blood volume, flow, or pressure~\cite{nichols2022mcdonald}. These dynamics can be inferred from a variety of observable signals, including electrical activity (electrocardiography, ECG)~\cite{goldberger2023goldberger}, optical absorption or transmission (photoplethysmography, PPG)~\cite{allen2007photoplethysmography}, Doppler frequency shifts (ultrasound)~\cite{hoskins2019diagnostic}, arterial pressure waveforms (tonometry)~\cite{nichols2022mcdonald}, flow-sensitive imaging (e.g., phase-contrast magnetic resonance imaging, PC-MRI)~\cite{mcrobbie2017mri}, and more recently, laser speckle contrast (speckle plethysmography, SPG)~\cite{boas2010laser}. While ECG is considered the clinical gold standard for cardiac sensing, PPG has emerged as the most prevalent modality, widely deployed across diverse form factors and applications, both clinical and consumer. Although both ECG and PPG enable accurate pulse rate estimation, they require physical contact with the subject. This is a particular limiting factor in scenarios such as neonatal care~\cite{aarts2013non}, infectious disease monitoring, burn patient assessment, and large-scale public health screening, where non-contact sensing is important.

\par Remote photoplethysmography (rPPG) addresses this need for non-contact sensing by enabling pulse rate estimation using imaging systems without direct contact and has been extensively studied, leading to algorithmic~\cite{verkruysse2008remote,kumar2015distanceppg,de2013robust,wang2016algorithmic,chen2018deepphys,yu2019remote,sun2022contrast,chari2024implicit} and dataset innovations~\cite{bobbia2019unsupervised,zhang2016multimodal,nowara2020near,joshi2024ibvp,vilesov2022blending,harish2025cogphys,tang2023mmpd}. However, rPPG relies on detecting subtle intensity variations in reflected light, making it inherently low in signal-to-noise ratio (SNR). As a result, its performance degrades significantly under changes in ambient illumination, subject or camera motion, and increasing imaging distance (see Supplementary Section 1.1 for more details on rPPG).

\par Laser speckle plethysmography, or speckle plethysmography (SPG), offers an alternative to PPG by exploiting temporal variations in laser speckle patterns. First introduced by Ghijsen \etal~\cite{ghijsen2018wearable}, SPG aims to record fluctuations in speckle signals during pulsatile cycles, making them less sensitive to absolute intensity. Coherent illumination (laser) produces speckle on rough surfaces through interference from microscopic surface and subsurface structures~\cite {boas2010laser,LeeRollingShutter24}. Pulsatile variations in blood flow modulate this speckle pattern over time, and the resulting fluctuations in speckle patterns encode the underlying hemodynamics~\cite{parker2024speckle,schnitzler2024wearable}. These fluctuations can be measured via spatial or temporal contrast, thus enabling robust estimation of vital signs such as heart rate (HR)~\cite{dunn2018comparison}, respiratory rate (RR)~\cite{hoogeveen2025three}, and blood pressure (BP)~\cite{lorato2024towards}. Similar to the evolution of PPG from contact to remote sensing, SPG  signals can also be estimated in a non-contact manner, i.e., remotely using cameras~\cite{olazabal2021beat}. Formally known as remote speckle plethysmography (rSPG), it inherits the robustness of its contact-based counterpart and has consistently outperformed rPPG-based systems~\cite{herranz2022camera, herranz2023comparison, herranz2023comparing}. However, most SPG studies are limited to contact or short-range scenarios, shown up to 25~cm~\cite{olazabal2021beat}. A parallel line of research into defocused speckle imaging, i.e., speckle vibrometry~\cite{zalevsky2009simultaneous,ozana2015demonstration,que2022contactless,zhu2024camera},  has been shown to work for larger distances, up to 60~m~\cite{zalevsky2009simultaneous} (see Supplementary Sections 1.2  and 1.3 for more details on SPG and Speckle Vibrometry).

\par Conventional rSPG uses frame-based intensity cameras to compute either spatial contrast per frame or temporal contrast over temporal windows, and then analyzes the resulting signal for pulsatile variations. We refer to these frame-based camera systems as intensity speckle plethysmography (iSPG) systems. However, these approaches are fundamentally limited by the dynamic range and frame rate of such cameras, resulting in degraded performance in low-light conditions, high ambient illumination, or at long distances. Moreover, in the presence of static speckle, the dynamic fluctuations observed using frame-based intensity cameras are reduced. Crucially, SPG is inherently driven by temporally varying contrast arising from rapid, low-amplitude fluctuations in speckle patterns. Event cameras~\cite{gallego2020event,lichtsteiner2008128,brandli2014240,posch2010qvga}, which asynchronously detect per-pixel intensity changes as \textit{events}, are therefore naturally suited to measure this signal. Their high temporal resolution ($\mu s$-scale latency) and high dynamic range ($>$ 120 dB) make them particularly well-suited for capturing fast, low-amplitude intensity variations. These properties have enabled a wide range of applications \cite{gallego2020event,chakravarthi2024recent} in robotics and autonomous driving~\cite{falanga2020dynamic,maqueda2018event,gehrig2024low}, 3D imaging~\cite{matsuda2015mc3d,muglikar2021esl,dashpute2026accurate}, microscopy~\cite{guo2024eventlfm,cabriel2023event}, and laser speckle imaging~\cite{cai2025event2audio,zhou2025event,ge2020dynamic,balaji2024probing}. Thus, they can effectively detect rapid fluctuations in speckle patterns, making them more resilient to ambient light, static speckle, and low signal-to-noise ratio, which is particularly important in challenging imaging conditions such as long-range imaging (see Supplementary Section S1.4 for more details on event cameras and speckle imaging).

\par In this work, we introduce event-based speckle plethysmography (eSPG), leveraging this natural match to directly measure temporal contrast with high dynamic range and temporal resolution, enabling robust pulse estimation across challenging imaging regimes. Through a series of stress-testing experiments, we demonstrate that rSPG, and particularly eSPG, remain accurate across an unusually wide range of conditions. Notably, eSPG achieves an error of 5 beats per minute (bpm) at a distance of $1~km$. The frame-based and event-based SPG imaging systems both maintain similar accuracy at short ($1~m$) and mid-range ($100~m$) distances, with eSPG outperforming in challenging motion conditions and when ambient, non-coherent light dominates the scene. We hypothesize that \textit{if you can resolve the speckle contrast reliably, you can measure the pulse rate}. In practice, this ability to resolve is a function of noise, magnification, speckle size, aperture, the sensor dynamic range, and turbulence. Thus, the choice of sensor and optics plays a pivotal role in resolving the speckle in extreme conditions such as long-distance sensing. Specifically, our choice of using an event camera, along with high-magnification optics, enables unreasonably effective remote HR sensing. We present and validate our findings through both theoretical design and experimental results. Accompanying our method is one of the largest datasets to date for SPG measurements, comprising videos from a frame camera, event streams from event sensors, and ground truth PPG signals from a pulse oximeter.

\par Our results indicate that eSPG enables robust, long-range, non-contact HR sensing across diverse real-world scenarios. Potential applications include public health monitoring, rescue operations, driver state sensing, and neonatal care, where contact-based sensing is often impractical. Our key contributions include:

\begin{itemize}
    \item \textbf{Event-based Speckle Plethysmography (eSPG):} We introduce eSPG, an improved low-power, high dynamic range, and fast SPG measurement modality, which involves imaging laser speckle patterns using an event camera. 
    
    \item \textbf{km-scale plethysmography:} We demonstrate accurate heart rate estimation at distances up to a km.
    
    \item \textbf{eSPG-derived HR and HRV estimation:} We propose methods and algorithms that enable accurate estimation of heart rate (HR) and heart rate variability (HRV) from eSPG waveforms.
    
    \item \textbf{Robustness of SPG:} We show that SPG is more robust than previously demonstrated in terms of ambient light, motion, and long-range distance imaging. We also introduce a robust frame-SPG method (iSPG++).
    
    \item \textbf{Public dataset and codebase:} We release one of the largest remote SPG datasets to date, and its accompanying codebase. The dataset contains 120 measurements at 1 m; 20 measurements each at 100 m, 350 m, 750 m, and 1 km; a total of 200 measurements, each 60 seconds long, for both frame and event, along with ground-truth PPG signals from a pulse oximeter.
\end{itemize}

\section{Results}
\label{sec:results}

\subsection{Principles}
\label{sec:principles}

\par We propose event-based speckle plethysmography (eSPG), a simple approach that illuminates the skin (e.g., palm) with a low-power red or NIR laser ($<5$~mW) and senses the resulting speckle dynamics using an event camera (\cref{fig:teaser}a,b). Temporal changes in the dynamic speckle pattern encode pulsatile blood flow and are directly captured as events by the event sensor. We show that the time-varying rate (or count) of events, as well as the spatial distribution of events, robustly and with high fidelity capture the underlying plethysmographic signal. This enables accurate pulse estimation under motion, varying illumination, and at significantly longer distances than prior methods (\cref{fig:teaser}c).
We further derive the operating conditions and theoretical bounds under which our method is effective. 

\subsection*{Speckle plethysmography (SPG)}
\label{sec:spg}

\vspace{5pt}
\par When coherent light (e.g., from a laser) is scattered by a disordered medium such as biological tissue, the multiple scattered wavefronts interfere to produce a granular intensity pattern known as speckle~\cite{goodman2007speckle}. While the tissue structure may be static, the motion of scatterers within pulsatile blood flow induces temporal fluctuations in the speckle pattern.
The speckle correlation time $\tau_c$ is roughly the time taken for a speckle pattern to change completely. Volumetric blood flow (SPG signal) is observed to be inversely proportional to $\tau_c$~\cite{ghijsen2018wearable,choi2006linear}. That is, faster volumetric flow results in a shorter speckle correlation time, and vice versa.

\begin{equation}
\text{SPG}(t) \propto 1/\tau_c,
\end{equation}

\par Similar to imaging a fast-moving object, rapid speckle fluctuations within the camera's exposure time are temporally integrated, resulting in a blurred speckle pattern and reduced speckle contrast. The statistical properties of the speckle pattern, such as its variance $\sigma^2$ and contrast $K$, can be analyzed to infer changes in blood flow and its pulsatile dynamics. The speckle contrast $K$ is usually calculated within a spatial or temporal window. If $\mu$ denotes the mean intensity of the signal in the window, and $\sigma$ denotes the standard deviation, $K$ is given by:
\begin{equation}
    K = \frac{\sigma}{\mu}.
\end{equation}

\noindent
The contrast $K$ is approximately related to $\tau_c$ as~\cite{boas2010laser}:
\begin{equation}
\tau_c \propto K^2,
\end{equation}

\noindent
leading to the (scaled) SPG signal being calculated as:
\begin{equation}
\text{SPG} = 1/K^2,
\label{eq:SPG_K}
\end{equation}

\subsection*{Event-based Speckle Plethysmography (eSPG)}
\label{sec:eSPG}

\vspace{5pt}
\par Sensing speckle fluctuations using frame-based intensity cameras has several limitations. The signal of interest (SPG waveform) is inherently a signal in the intensity gradient domain. Speckle fluctuations are measured by measuring the \textit{contrast} in the recorded speckle intensity. Event cameras, which naturally sense \textit{temporal intensity gradients} at native temporal resolution, are thus better suited for sensing speckle pattern fluctuations. Event cameras directly detect intensity fluctuations as asynchronous events and respond to changes in log intensity. Thus, they naturally end up suppressing static scattering components, ignoring contributions from ambient non-coherent light, and directly capturing temporal speckle fluctuations. Their high dynamic range further enables reliable detection of contrast changes even in low-light conditions. As a result, the measured event activity is closely related to the underlying speckle decorrelation dynamics, and hence to the speckle correlation time $\tau_c$.

\par An \textit{event} $e$ is triggered when the change in the log intensity (I) crosses a certain set threshold $\epsilon$ in a certain time $\Delta t$. For each such event, the camera reads out the tuple of the location of its pixel ($x, y$), its corresponding timestamp $t$, and its polarity $p$, 

\begin{equation}
    e_{i} = (x_i, y_i, t_i, p_i),
\end{equation}
such that,
\begin{equation}
    | \log(I(x, y, t)) - \log(I(x,y,t-\Delta t)) | > \epsilon,
\end{equation}
where $p$ indicates the polarity of the intensity change.

\vspace{5pt}
\par Based on events arising from temporally varying speckle fluctuations, we propose generating event-based SPG waveforms in two ways, as listed below. 

\vspace{5pt}
\noindent
\textbf{Event count waveform.} We consider a scenario where a laser illuminates a spot on human skin tissue. When imaging such speckle with an event camera, we observe events in and around the laser illumination spot, due to random fluctuations in the speckle pattern. These fluctuations and changes occur due to the motion of blood in the arteries beneath the skin. We observe that both the total number of events and the spatial distribution of events around the laser illumination spot change with pulsatile blood flow (see \cref{fig:teaser}d,e). Consider an event stream $\{e_i\}_{i=1}^M$ divided into uniformly-spaced time-bins ${t_1, t_2, ..., t_K}$. The number of events occurring within a time bin is given by 

\begin{equation}
    N_k = \sum_i \textbf{1} \{t_i \in (t_{k-1}, t_k] \}.
\end{equation}

\noindent
\textbf{Radial spread waveform.} Consider the illumination spot to be centered at $(x_c,y_c)$ in the event camera pixel coordinates. For every event $i$, \textit{radial distance} of the event from the illumination center is computed as $r = \sqrt{(x_i-x_c)^2+(y_i-y_c)^2)}$. Let ${r_1, r_2, ..., r_D}$ correspond to radial distances (in pixels) from the laser illumination spot. For every time-bin $k$, we create a \textit{radial histogram} of events given by 

\begin{equation}
    N_k(r_d) = \sum_i \textbf{1} \{t_i \in (t_{k-1}, t_k],  r \in (r_{d-1}, r_d]\}, 
\end{equation}

\par We normalize the radial histogram of event counts to obtain the radial distribution of events for every time-bin $k$, which is given by $\rho_k(r_d)$:   

\begin{equation}
    \rho_k(r_d) = N_k(r_d)/ \sum_d N_k(r_d).
\end{equation}

\par Then we obtain an \textit{effective event radius}, $r_k^{eff}$, by weighting each radial zone by the radial distribution of events as follows: 

\begin{equation}
    r_k^{eff} = \sum_j r_j.\rho_k(r_j).
    \label{eq:r_effective}
\end{equation}

\par We plot the time series generated by $r_k^{eff}$ and $N_k$ for all $t_k$ to obtain our plethysmograph (eSPG) waveforms as shown in \Cref{fig:teaser}d,e.

\subsection*{Baseline plethysmography algorithms}

\vspace{5pt}
\par We compare our eSPG waveforms with the ground-truth PPG signal obtained from a pulse oximeter, the state-of-the-art remote PPG algorithm, distPPG~\cite{kumar2015distanceppg}, and baseline remote SPG waveforms generated from intensity (frame-based) cameras, whose algorithms we outline below.

\vspace{5pt}
\noindent
\textbf{Intensity-based SPG (iSPG) algorithm:} Based on \cite{olazabal2021beat}, we compute the speckle contrast ($K$) as 
\begin{equation}
    K = \frac{1}{N}\sum_{i} \frac{<\sigma>_{\Omega_i^{7\times7}}}{<\mu>_{\Omega_i^{7\times7}}}
\end{equation}
where $<\sigma>_{\Omega_i^{7\times7}}, <\mu>_{\Omega_i^{7\times7}}$ are standard deviation and mean calculation operations done over a $7\times7$ window at location $i$. Averaging over all such windows with laser illumination yields the final speckle contrast estimate. This speckle contrast estimate is computed across time to give the time-varying iSPG waveform: $$iSPG(t) = 1 / K(t)^2$$

\vspace{5pt}
\noindent
\textbf{Proposed iSPG++ algorithm (Ours):} Taking inspiration from~\cite{kumar2015distanceppg}, we employ the \textit{Goodness metric} with the spatial windowing logic to take a weighted average of the spatial windows to modify the iSPG calculation. We weight the contrast waveform from each region by its quality metric before fusion. Mathematically, this can be summarized as: 
\begin{align}
\text{iSPG++}(t) = \sum_{i=1}^N G_i\cdot\text{K}_i(t),
\end{align}

\noindent
where $\text{K}_i$ is the contrast signal for the $i^{th}$ sliding window, and $G_i$ is the Goodness metric from~\cite{kumar2015distanceppg} modified to include the energy of the first harmonic of the pulse rate.
% \ad{add a line about distPPG also so we have talked about all the baseline methods...something like "Additionally, we also compare to the state-of-the-art remote PPG method distPPG~\cite{kumar2015distanceppg}."}

% \begin{align}
% \ G_i(PR) = \frac{\int_{PR-b}^{PR+b} \hat{Y}_i(f)\mathrm{d}f}{\int_{B_1}^{B_2} \hat{Y}_i(f)\mathrm{d}f - \int_{PR-b}^{PR+b} \hat{Y}_i(f)\mathrm{d}f}.
% \end{align}

\subsection{Experimental Details}
\label{sec:experiments}

\vspace{10pt}

\par We evaluate our method on human participants across a variety of scenarios, each examining a different dimension of robustness. Our first experiment tests the efficacy of our imaging setup at long distances. Specifically, we image at $100~m$, $350~m$, $750~m$ and $1~km$, enabling imaging distances that are an order of magnitude longer than prior work~\cite{blackford2017measurements,blackford2016long}. Following the evaluation at long distances, we stress-test our laser speckle imaging setup at a typical short operating distance ($<1~m$) under different degradations: hand motion, laser motion, changes in laser profile and wavelength, and extreme illumination.

\par We choose the participant's hand (palm) as our region of interest (ROI) across all regimes. That is, the participant's palm is illuminated and imaged. The exact ROI used for signal computation is manually selected once per recording. The core sensing pipeline, comprising laser-based speckle generation and acquisition, remains consistent across all experiments. We split the experiments in three regimes based on their imaging distance: (1) long range measurements, where the imaging distances are $350~m$, $750~m$ and $1~km$ while the laser illumination distance is $60~m$ from the participant because of laser safety and laser beam divergence issues, (2) mid range measurements, where both the laser and projector are each $100~m$ from the participant, and (3) short range measurements, where the laser-projector setup is placed $<1~m$ from the participant. We first describe the common experimental setup shared across all three regimes, followed by a detailed account of the modifications made for each distance. 

% \subsection{Common Experimental Setup} \label{sec:experimental_setup}

\vspace{10pt}
\noindent \textbf{Common Experimental Setup.} We keep the same frame- and event-based sensors across different experiments. We use a \textit{Ximea MC124MG-SY} monochrome camera for frame-based speckle measurements, such as iSPG~\cite{olazabal2021beat}, and absorption measurements such as distancePPG~\cite{kumar2015distanceppg}. Further, we employ a \textit{Prophesee EVK Gen 4} event camera to capture the eSPG signal. The eSPG signal can be computed from one of two features: the event count or the radial spread of the events, as shown in \cref{fig:teaser}d,e. \Cref{sec:choice_eSPG_feature} describes a principled approach to choosing the right feature (event count or radial spread) based on radial spread fluctuations.

\par A laser illuminates the participant's palm, which serves as our imaging target. The resulting subsurface scattering from the palm is then captured. The choice of imaging optics for each set of experiments is based on the required imaging distance and magnification, and is detailed in the following subsections. The optics must also satisfy a trade-off between speckle size and signal-to-noise ratio (SNR) to ensure speckles are sufficiently resolved at the sensor while maintaining adequate light collection. See \cref{sec:long_range_tradeoff} for a detailed explanation and Supplementary Sections S2.1 and S2.2 for all the calculations.

\par To validate the effectiveness of our sensing stack in predicting HR, we include a contact-based fingertip pulse oximeter \textit{Contec CMS60C} to record ground-truth PPG signals. In addition to the raw data, we collect timestamps from each sensor to enable effective comparisons and ensure coarse temporal synchronization. All our experiments were conducted with Institutional Review Board protocols (IRB; study number: IRB-FY2026-159) approved by Rice University. 

% \subsection{Metrics} \label{sec:metrics}
\vspace{10pt}
\noindent \textbf{Metrics.} Across all experiments, we compute the following metrics to quantify and characterize the performance: mean absolute error (MAE), root mean squared error (RMSE), mean absolute percentage error (MAPE), and Pearson correlation coefficient (r) - all of which are computed on the HR values themselves. We also report the clinical significance score (AAMI)~\cite{association2002cardiac} for each set of measurements. Formal definitions of each metric are reported in Methods (\cref{sec:algo_metrics}). Additionally, we also provide the Bland-Altman plots for all experiments in the Supplementary Sec. S3.4.

\subsection{Long range: Plethysmography at 1~km}

\begin{figure}[!ht]
    \centering
    \includegraphics[width=\linewidth]{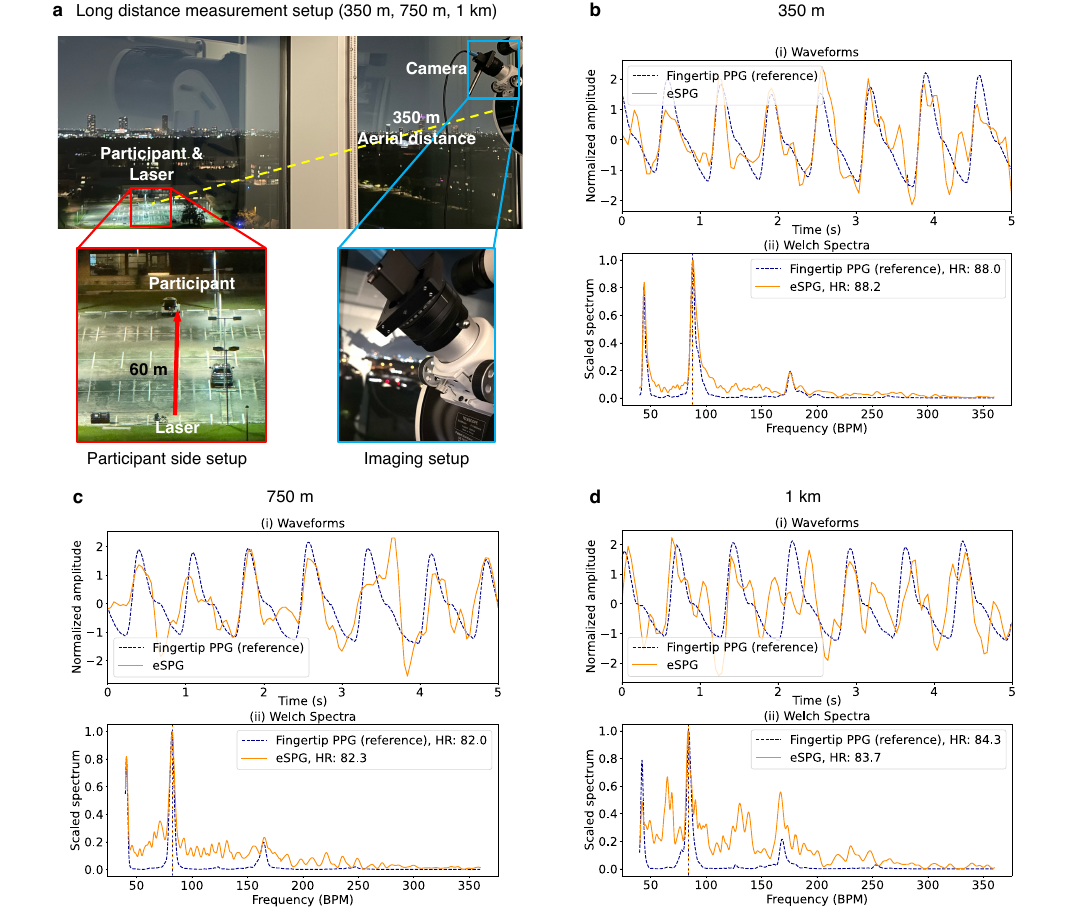}
    \caption{\textbf{\boldmath eSPG signals can be imaged with high fidelity even at long imaging distances ($350~m, 750~m, 1~km$).} \textbf{a,} Our experimental setup consists of a camera (event camera for eSPG or frame camera for iSPG) mounted on a telescope on the $10^{th}$ floor of a building. Here, we illustrate the relative scale of the participant in the car with respect to the imaging distance - $350~m$ in this case. The participant is seated in a car, and the event camera and frame-based camera are alternated after each recording. The laser (operated at 4.99 mW) is placed $\approx60~m$ from the participant, ensuring adequate beam coverage of the hand without significant power loss from its divergence. We capture measurements at three distances: \textbf{b} $350~m$, \textbf{c,} $750~m$, and \textbf{d,} $1~km$. Across all three distances, the periodic nature of the plethysmography signals is captured with high fidelity. While SNR degrades at greater distances, the dominant frequency peak is still discernible at $1~km$.}
    \label{fig:experimental_setup_long}
\end{figure}

% \subsubsection{Experimental Setup}

\vspace{10pt}
\noindent\textbf{Experimental Setup.} \Cref{fig:experimental_setup_long}a illustrates the experimental setup for long-range experiments, with the participant located at $350~m$. The same setup is also used to image participants at $750~m$ and $1~km$. Per the constraints elaborated in \cref{sec:long_range_tradeoff}, long-range imaging requires a high-magnification optical setup. Consequently, we employ a \textit{Sky-Watcher 300P} telescope ($f/5$, focal length $f$ = $1500~mm$, aperture $D$ = $300~mm$) as the sensing optic. We alternate measurements between the event- and frame-based cameras to maximize light throughput, thereby avoiding the need for a beam splitter. During imaging, the cameras were equipped with a $633~nm$ narrow-bandpass filter (\textit{Thorlabs FLH633-5}). The participant's palm is illuminated by a helium-neon (HeNe) laser (\textit{LASOS LGK 7628}) operating at $\lesssim 5~mW$, located $\approx 60~m$ away. We choose this distance to adhere to laser-safety constraints in public environments and alleviate beam divergence limitations that make $350~m - 1~km$ illumination impractical without specialized beam shaping. We recruit 5 participants for each imaging distance and capture 4 trials per participant per camera (twice for each hand). This gives us a total of $60~mins$ of recordings for each camera. The participants remain stationary throughout the recordings. 

% \subsubsection{Results}

\vspace{10pt}
\noindent\textbf{Results.} The qualitative waveforms are shown in \cref{fig:experimental_setup_long}b,c,d and the quantitative results are summarized in \cref{tab:long_dist}. Across the 3 distances, eSPG is the most stable and consistently performing method. On the other hand, frame-based methods such as iSPG~\cite{olazabal2021beat} and iSPG++ (ours) perform better at $350~m$ but deteriorate quickly at longer distances. A closer look at $350~m$ reveals the stability of eSPG: despite having a higher MAE than iSPG++, the event modality has just as many samples that clear the clinical test (AAMI). This reveals that eSPG produced a couple of estimates with high errors, skewing the MAE. The RMSE further corroborates this: iSPG++ and eSPG both have similar MAEs, but eSPG has a higher RMSE. This indicates that eSPG had a few more extreme samples.

\par At longer distances - $750~m$ and $1~km$ - the performance difference between the event and frame-based cameras is apparent. The AAMI clinical metric remains nearly constant, and the drop in MAE is much smaller than that observed with frame-based methods. Thus, the combination of these metrics leads us to believe that while increasing distances reduce the estimation quality for all methods, event cameras are more resilient. That is, eSPG has consistently yielded estimates that meet the clinical threshold, with only a couple of samples failing the test due to high error. 

\begin{table*}[!ht]
\centering
\caption{\textbf{\boldmath
Long-range: eSPG shows consistently high performance across across 3 long-range imaging distances ($350~m, 750~m, 1~km$).} For each of the three imaging distances, the data were collected by illuminating the palms of 5 participants, with 4 measurements per participant (two samples from each hand). As the imaging distances increase, the performance of the intensity-based methods degrades. This is particularly evident in the simultaneous rise in MAE and the drop in AAMI clinical threshold score. On the other hand, the event-based method is more stable, as evidenced by a consistent MAE and AAMI score across the 3 imaging distances.
}
\begin{tabular}{c l c c c c c}
\toprule
 & & \multicolumn{4}{c}{HR Error ($BPM$)}&\multicolumn{1}{c}{Clinical}\\
\cmidrule(lr){3-6} \cmidrule(lr){7-7}
 & Method & MAE $\downarrow$ & RMSE $\downarrow$ & MAPE(\%) $\downarrow$ & $r$ $\uparrow$ & AAMI $\uparrow$ \\
\midrule
\multirow{4}{*}{\rotatebox{90}{\textbf{\boldmath$350~m$}}} & distPPG & $52.2 \pm 4.7$ & $56.3 \pm 20.3$ & $63.0 \pm 6.3$ & $0.24 \pm 0.23$ & \runnerup{$1/20$} \\
 & iSPG & \runnerup{$3.4 \pm 1.7$} & \runnerup{$8.5 \pm 7.8$} & \runnerup{$4.1 \pm 2.1$} & \runnerup{$0.65 \pm 0.18$} & \winner{$18/20$} \\
 & \textbf{iSPG++} (ours) & \winner{$1.8 \pm 0.6$} & \winner{$3.2 \pm 2.4$} & \winner{$2.2 \pm 0.7$} & \winner{$0.93 \pm 0.09$} & \winner{$18/20$} \\
\cmidrule(lr){2-7}
 & \textbf{eSPG} (ours) & $3.6 \pm 2.2$ & $10.4 \pm 9.5$ & $4.3 \pm 2.6$ & $0.32 \pm 0.22$ & \winner{$18/20$} \\
\midrule
\multirow{4}{*}{\rotatebox{90}{\textbf{\boldmath$750~m$}}} & distPPG & $24.7 \pm 3.1$ & $28.3 \pm 12.9$ & $29.4 \pm 3.7$ & $-0.42 \pm 0.21$ & $4/20$ \\
 & iSPG & \runnerup{$7.6 \pm 2.9$} & \runnerup{$15.0 \pm 10.8$} & \runnerup{$8.9 \pm 3.3$} & \runnerup{$0.24 \pm 0.23$} & \runnerup{$15/20$} \\
 & \textbf{iSPG++} (ours) & $8.1 \pm 3.8$ & $18.7 \pm 16.1$ & $10.3 \pm 5.0$ & $0.09 \pm 0.23$ & \winner{$17/20$} \\
\cmidrule(lr){2-7}
 & \textbf{eSPG} (ours) & \winner{$4.0 \pm 1.8$} & \winner{$8.8 \pm 7.0$} & \winner{$4.6 \pm 2.0$} & \winner{$0.61 \pm 0.19$} & \winner{$17/20$} \\
\midrule
\multirow{4}{*}{\rotatebox{90}{\textbf{\boldmath$1~km$}}} & distPPG & $25.0 \pm 4.5$ & $32.1 \pm 17.7$ & $32.3 \pm 6.6$ & $-0.17 \pm 0.23$ & $5/20$ \\
 & iSPG & \runnerup{$8.1 \pm 3.0$} & \runnerup{$15.6 \pm 11.9$} & \runnerup{$11.3 \pm 4.5$} & $0.09 \pm 0.23$ & \runnerup{$16/20$} \\
 & \textbf{iSPG++} (ours) & $11.2 \pm 3.0$ & $17.5 \pm 11.6$ & $14.9 \pm 4.2$ & \runnerup{$0.22 \pm 0.23$} & $13/20$ \\
\cmidrule(lr){2-7}
 & \textbf{eSPG} (ours) & \winner{$5.0 \pm 3.0$} & \winner{$14.4 \pm 12.4$} & \winner{$6.0 \pm 3.7$} & \winner{$0.49 \pm 0.21$} & \winner{$18/20$} \\
\bottomrule
\end{tabular}
\label{tab:long_dist}
\end{table*}

\subsection{Mid range: 100~m experiments}

\begin{figure}[!ht]
    \centering
    \includegraphics[width=\linewidth]{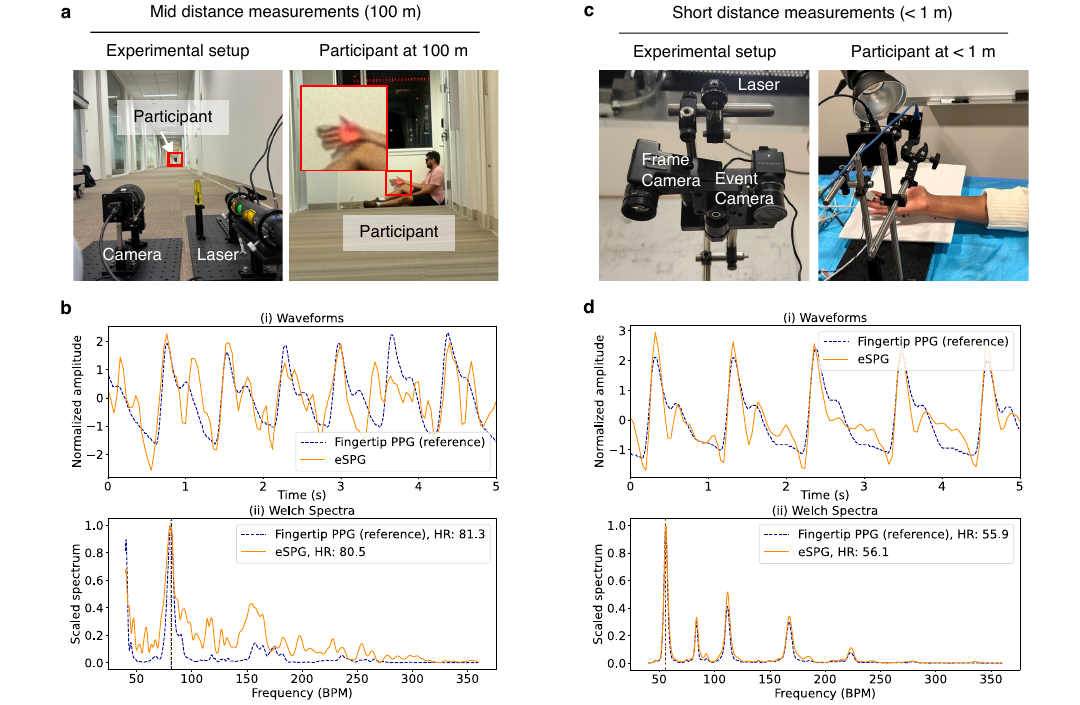}
    \caption{\textbf{\boldmath eSPG reliably captures HR at $100~m$ laser and camera distance and is robust to motion and extreme illumination at shorter distances ($<1~m$).} \textbf{a,} The participant is imaged $100~m$ from the laser-camera setup. The cameras are alternated between trials, and the ground-truth signals are captured using a pulse oximeter on the participant's hand. \textbf{b,} eSPG signals are faithfully estimated from the participants for the $100~m$ setup. \textbf{c,} For the short-range ($<1~m$) measurements, both cameras and the laser are mounted together as a single stack, and recordings are captured under various operating conditions: motion, varied laser profiles, and extreme ambient illumination. \textbf{d,} At such short distances, we see the faithful capture of higher-order harmonics, especially when the hand is kept stationary. A stratified analysis of the various imaging conditions is presented in \cref{tab:1m_all}.}
    \label{fig:experimental_setup_short_mid}
\end{figure}

% \subsubsection{Experimental Setup.}
\vspace{10pt}
\noindent\textbf{Experimental Setup.} \Cref{fig:experimental_setup_short_mid}a depicts our experimental setup for mid-range experiments. Here, we position the laser and camera setup at $\approx100~m$ from the participant. Specifically, the laser and the camera are placed adjacent to each other, rather than at different distances. We equip the camera with a \textit{Sigma lens} ($f=135~mm$) to achieve the required magnification for speckle imaging, and alternate between the event- and frame-based cameras. Based on the signal-speckle size trade-offs discussed in \cref{sec:long_range_tradeoff}, we set the aperture of the lens to its full size ($f/1.8$) to maximize signal collection in the low-light regime. Reducing the aperture further degrades the signal strength more than supporting the increasing speckle size. The participant is seated while the laser illuminates the palm, with the beam spot naturally expanded due to propagation-induced divergence. A 2mW HeNe laser (\textit{Thorlabs HRP020-1}) was used for these trials. The dataset comprises 20 recordings per sensor from 5 participants, with each participant contributing 4 recordings per sensor (twice per hand). The participants remain stationary throughout the recordings.

\vspace{10pt}
\noindent\textbf{Results.} As seen in \cref{tab:hr_100m}, the proposed eSPG outperforms the best frame-based SPG methods by almost $2\times$ in HR estimation errors. This consequently leads to eSPG meeting the clinical accuracy criteria for all measurements across all participants.  This is further evidenced by the clear temporal and frequency alignments in the example waveforms in \cref{fig:experimental_setup_short_mid}b. In addition to eSPG, our other proposed method, iSPG++, emerges as the second-best overall and best frame-based approach at $100~m$ imaging, outperforming the previous baseline by a factor of $2\times$. Whilst iSPG++ achieves a low MAE of $2~BPM$, three samples failed the clinical metrics.

\begin{table*}[!ht]
\centering
\caption{\textbf{\boldmath All SPG methods, particularly eSPG, show promising performance on mid-range imaging, i.e., imaging and illumination distances at $100~m$.} The data was collected by illuminating the palms of 5 participants, with 4 measurements per participant (two samples from each hand).
Across all samples, the event-based method outperforms all intensity baselines, with eSPG estimates passing the AAMI clinical threshold for all 20 samples.}
\begin{tabular}{l c c c c c}
\toprule
 & \multicolumn{4}{c}{HR Error ($BPM$)}&\multicolumn{1}{c}{Clinical}\\
\cmidrule(lr){2-5} \cmidrule(lr){6-6}
Method & MAE $\downarrow$ & RMSE $\downarrow$ & MAPE(\%) $\downarrow$ & $r$ $\uparrow$ & AAMI $\uparrow$ \\
\midrule
distPPG & $68.8 \pm 3.0$ & $70.1 \pm 20.6$ & $90.3 \pm 7.8$ & $0.02 \pm 0.24$ & $0/20$ \\
iSPG & $4.1 \pm 1.7$ & $8.6 \pm 7.5$ & $5.8 \pm 2.8$ & $0.77 \pm 0.15$ & $17/20$ \\
\textbf{iSPG++} (ours) & \runnerup{$2.0 \pm 0.6$} & \runnerup{$3.5 \pm 2.5$} & \runnerup{$2.5 \pm 0.8$} & \runnerup{$0.97 \pm 0.06$} & \runnerup{$19/20$} \\
\midrule
\textbf{eSPG} (ours) & \winner{$1.1 \pm 0.2$} & \winner{$1.4 \pm 0.8$} & \winner{$1.3 \pm 0.2$} & \winner{$1.00 \pm 0.02$} & \winner{$20/20$} \\
\bottomrule
\end{tabular}
\label{tab:hr_100m}
\end{table*}

\subsection{Short range: <1~m experiments}

% \subsubsection{Experimental Setup.}

\vspace{10pt}
\noindent \textbf{Experimental Setup.} \Cref{fig:experimental_setup_short_mid}c shows the experimental setup for short-range experiments. Additional setups are shown in Supplementary Sec. 2.3 (Fig. S1). Unlike the previous experiments, both event- and frame-based cameras are mounted together, and the participant, located at $\approx 0.6~m$, is imaged by both cameras simultaneously. Both cameras are equipped with a machine vision lens, \textit{Fujinon $12.5~mm$ lens (HF12.5HA-1B)}, and the aperture is set at $f/6$ to satisfy the speckle-size-SNR constraints mentioned in  \cref{sec:long_range_tradeoff}. For this setup, we recruited 10 participants and evaluated the method across five experiments, each systematically assessing its robustness along various dimensions.
% obfuscation usage is incorrect here.
Specifically, we test the effect of changes in laser wavelength \& profile, extreme ambient light, participant motion, and laser motion.\\

\par\noindent\textbf{A. Still:} We image the participant's palm under controlled, stationary conditions. A fixed $850~nm$ near-infrared (NIR) laser (\textit{Thorlabs CPS850S} at 3-4~$mW$) is used as the illumination source.\\

\par\noindent\textbf{B. Effect of Illumination Sensitivity :} We image the participant's palm area under controlled, stationary conditions with varying laser illumination configurations in ambient room light. The varying illumination conditions are listed as follows: 

\begin{enumerate}
    \item Gaussian profile: (Helium-Neon) HeNe laser beam at $633~nm$ and $\lesssim 2~mW$ power.
    \item Uniform profile: HeNe laser beam with optical beam expansion.
    \item Sub-$mW$ laser: Classroom laser pointer ($0.5~mW$ laser).
\end{enumerate}

\noindent\textbf{C. Effect of Ambient Light:} We record the participant's stationary palm under two levels of ambient lighting, each more than four orders of magnitude higher than the laser illumination power ($<4~mW$). We flood-illuminate the scene with a halogen lamp (Godox SL-200W) at two levels:
\begin{enumerate}
    \item High ambient light at $80~W$,
    \item Very high ambient light at $160~W$.
\end{enumerate}

\noindent\textbf{D. Effect of Subject (Hand) Motion:} We capture video recordings of the participant's palm illuminated by the Thorlabs NIR laser. The participant is asked to move their hand in various ways to test the system's robustness. The types of motion are listed as follows: 
\begin{enumerate}
    \item Small motion: The palm hovers in the air, with the tremors inducing small motion
    \item Large motion: The hand moves exaggeratedly, while ensuring the laser spot remains on the palm.
\end{enumerate}

\noindent\textbf{E. Effect of Laser Motion:} We capture video recordings of the palm illuminated by the handheld sub-$mW$ classroom laser pointer. Motion is induced by natural hand tremors when holding the laser. The specific cases are:
\begin{enumerate}
    \item Laser motion: Handheld sub-$mW$ laser illuminating a stationary palm.
    \item Dual (laser-subject) motion: Handheld sub-$mW$ laser illuminating a palm hovering in the air, i.e., with small motions.
\end{enumerate}

\vspace{10pt}
\noindent \textbf{Results.} \Cref{tab:1m_all} summarizes the results from stress-testing our imaging setup at short distances of $<1~m$. In the simplest case (still samples at $850~nm$ illumination wavelength), speckle-based methods perform well with both event and frame-based cameras, as they operate in a favorable regime of background light, speckle size, and signal-to-noise ratio. Further, our imaging setup is also observed to be robust to laser illumination changes, with all three beam profiles at $633~nm$ - uniform, Gaussian, and sub-$mW$ - together achieving an average HR estimation error $<1~BPM$. This robustness is not limited to event cameras and extends to frame-based cameras as well. Hence, we experimentally show that the choice of laser itself, even when extreme (sub-$mW$), has a low impact on HR estimation performance.

\par The presence of strong ambient light severely reduces visible speckle by saturating the frame-based camera, hence failing to work for it in most of the measurements. The high dynamic range of event cameras aids sensing under such harsh illumination, making eSPG a much more robust modality.

\par For participant hand motion, eSPG was the best-performing method, achieving both the lowest HR error and the highest clinical significance. Furthermore, eSPG is the only method that achieves an average error below $5~BPM$. When there is relative motion between the event camera and the laser (e.g., due to laser motion), our method performs poorly and underperforms the baselines, which is a limitation of our method.
% We believe that the radial features for eSPG alone might not be robust to laser motion and may require algorithmic changes to improve performance.

% On the other hand, we find the eSPG is not as effective as frame-based methods such as iSPG. While the RMSE was the lowest, the drop in clinical significance metrics was stark. However, we also note that iSPG++, a frame-based algorithm we proposed, yields results close to the baseline iSPG. The HR errors are observed to be within $0.5~BPM$ of the baseline, and only 1 extra sample failed the clinical significance test. Hence, we believe that the radial features for eSPG alone might not be robust to laser motion and may need alternate approaches to improve performance.

\par In addition to these five sets of experiments at the short range, we conduct two additional experiments: (a) dorsal imaging of the hand, and (b) vascular occlusion, to validate whether our method captures physiologically meaningful changes. These are elaborated on in the Supplementary Sec. S3.1 (Table S1), and Sec. S3.2 (Fig. S2), respectively.

\begin{table*}[ht]
\centering
\caption{\textbf{\boldmath eSPG demonstrates robustness to external factors such as corruption from motion, ambient illumination, and changes to the laser power and beam profile.} The \textbf{still} case is the baseline reading where the participant remains stationary. On more complex tasks, such as \textbf{hand motion} and extreme \textbf{ambient light}, eSPG outperforms all frame-based methods. Further, eSPG is not sensitive to the choice of laser illumination, with our method yielding low error for sub-$mW$ lasers and for HeNe lasers with Gaussian and uniform beam profiles. However, our method is not robust to the laser's motion. In this case, eSPG produces higher errors than its baselines, which is a limitation of our method.}
\begin{tabular}{c l c c c c c}
\toprule
 & & \multicolumn{4}{c}{HR Error ($BPM$)}&\multicolumn{1}{c}{Clinical}\\
\cmidrule(lr){3-6} \cmidrule(lr){7-7}
 & Method & MAE $\downarrow$ & RMSE $\downarrow$ & MAPE(\%) $\downarrow$ & $r$ $\uparrow$ & AAMI $\uparrow$ \\
\midrule
\multirow{4}{*}{\rotatebox{90}{\textbf{Still}}} & distPPG & $24.6 \pm 4.9$ & $29.0 \pm 17.7$ & $37.7 \pm 9.0$ & $-0.20 \pm 0.35$ & $1/10$ \\
 & iSPG & \winner{$0.7 \pm 0.5$} & \winner{$1.8 \pm 1.8$} & \winner{$0.9 \pm 0.6$} & \winner{$0.99 \pm 0.05$} & \winner{$10/10$} \\
 & \textbf{iSPG++} (ours) & \runnerup{$3.9 \pm 3.0$} & \runnerup{$10.4 \pm 10.0$} & $5.3 \pm 4.2$ & \runnerup{$0.71 \pm 0.25$} & \runnerup{$9/10$} \\
\cmidrule(lr){2-7}
 & \textbf{eSPG} (ours) & \winner{$0.7 \pm 0.5$} & \winner{$1.8 \pm 1.8$} & \runnerup{$1.0 \pm 0.6$} & \winner{$0.99 \pm 0.05$} & \winner{$10/10$} \\
\midrule
\multirow{4}{*}{\rotatebox{90}{\textbf{\shortstack{Hand\\Motion}}}} & distPPG & $22.8 \pm 4.3$ & $30.0 \pm 16.6$ & $30.8 \pm 5.8$ & $-0.10 \pm 0.23$ & $4/20$ \\
 & iSPG & \runnerup{$5.2 \pm 3.0$} & \runnerup{$14.4 \pm 12.7$} & \runnerup{$6.8 \pm 3.7$} & \winner{$0.72 \pm 0.16$} & \winner{$18/20$} \\
 & \textbf{iSPG++} (ours) & $5.8 \pm 3.0$ & $14.6 \pm 12.7$ & $7.6 \pm 3.7$ & \runnerup{$0.70 \pm 0.17$} & \runnerup{$17/20$} \\
\cmidrule(lr){2-7}
 & \textbf{eSPG} (ours) & \winner{$3.7 \pm 2.0$} & \winner{$9.9 \pm 8.9$} & \winner{$4.8 \pm 2.5$} & $0.62 \pm 0.19$ & \winner{$18/20$} \\
\midrule
\multirow{4}{*}{\rotatebox{90}{\textbf{\shortstack{Ambient\\Light}}}} & distPPG & $73.9 \pm 4.1$ & $76.2 \pm 21.9$ & $107.4 \pm 8.4$ & $0.04 \pm 0.24$ & $1/20$ \\
 & iSPG & \runnerup{$59.4 \pm 6.9$} & \runnerup{$67.0 \pm 25.4$} & \runnerup{$85.6 \pm 10.9$} & \runnerup{$0.14 \pm 0.23$} & $3/20$ \\
 & \textbf{iSPG++} (ours) & $60.9 \pm 7.1$ & $68.8 \pm 25.4$ & $88.7 \pm 11.7$ & $0.10 \pm 0.23$ & \runnerup{$4/20$} \\
\cmidrule(lr){2-7}
 & \textbf{eSPG} (ours) & \winner{$5.0 \pm 2.3$} & \winner{$11.6 \pm 9.8$} & \winner{$6.3 \pm 2.8$} & \winner{$0.62 \pm 0.19$} & \winner{$16/20$} \\
\midrule
\multirow{4}{*}{\rotatebox{90}{\textbf{\shortstack{Illumination\\Sensitivity}}}} & distPPG & $18.5 \pm 4.0$ & $28.6 \pm 16.8$ & $27.9 \pm 5.9$ & $0.24 \pm 0.18$ & $12/30$ \\
 & iSPG & \runnerup{$1.0 \pm 0.3$} & $2.1 \pm 1.6$ & $1.5 \pm 0.5$ & \runnerup{$0.98 \pm 0.04$} & \runnerup{$29/30$} \\
 & \textbf{iSPG++} (ours) & \runnerup{$1.0 \pm 0.3$} & \runnerup{$2.0 \pm 1.6$} & \runnerup{$1.4 \pm 0.5$} & \runnerup{$0.98 \pm 0.04$} & \runnerup{$29/30$} \\
\cmidrule(lr){2-7}
 & \textbf{eSPG} (ours) & \winner{$0.8 \pm 0.2$} & \winner{$1.2 \pm 0.8$} & \winner{$1.0 \pm 0.2$} & \winner{$0.99 \pm 0.02$} & \winner{$30/30$} \\
\midrule
\multirow{4}{*}{\rotatebox{90}{\textbf{\shortstack{Laser\\Motion}}}} & distPPG & $19.9 \pm 2.7$ & $23.3 \pm 10.6$ & $27.0 \pm 3.3$ & $-0.13 \pm 0.23$ & $3/20$ \\
 & iSPG & \winner{$3.8 \pm 2.0$} & \winner{$9.8 \pm 8.7$} & \winner{$5.1 \pm 2.6$} & \runnerup{$0.59 \pm 0.19$} & \winner{$17/20$} \\
 & \textbf{iSPG++} (ours) & \runnerup{$4.2 \pm 2.0$} & \runnerup{$10.0 \pm 8.7$} & \runnerup{$5.7 \pm 2.6$} & \winner{$0.64 \pm 0.18$} & \winner{$17/20$} \\
\cmidrule(lr){2-7}
 & \textbf{eSPG} (ours) & $6.5 \pm 2.1$ & $11.5 \pm 8.2$ & $8.8 \pm 2.7$ & $0.55 \pm 0.20$ & \runnerup{$13/20$} \\
\bottomrule
\end{tabular}
\label{tab:1m_all}
\end{table*}

\section{Discussion}
\label{sec:discussion}
In this work, we introduced event camera-based remote speckle plethysmography to reliably and robustly estimate vital signs. Event cameras are uniquely suited to detect temporal changes in intensity, and we leveraged this to detect speckle fluctuations in human tissue caused by underlying blood flow. By imaging at long-range distances (up to 1 km), we demonstrated the robustness of eSPG-based HR estimation. Further experiments at close range (< 1 m) demonstrated accurate HR estimation under challenging conditions, such as subject motion and high ambient lighting. We also demonstrated accurate HRV and respiratory rate estimates from eSPG waveforms at short range (Supplementary Sec. 3.3). 

% Limitations and Future work
\par Our method shows limitations when there is relative motion between the camera and the illumination. Additionally, our computations involve interactive region-of-interest (ROI) selection (for all methods, both proposed and baseline), which may affect the resulting HR values. Automated, intelligent ROI selection for an even more robust eSPG, alongside machine-learning-based estimation models, is a promising direction for future work. To utilize the full temporal resolution properties, spiking neural networks can also be used as part of the HR waveform generation algorithm for eSPG. With more robust algorithms, higher laser power, and appropriate optics, we envision this method working for even larger imaging distances in the future. Future work can also include more extensive clinical validation and feasibility studies for specific applications like neonatal care, burn patient monitoring, and disaster rescue operations.

\section*{Acknowledgments}
\label{sec:acknowledgments}

This work is supported by NIH DeepDOF - R01DE032051 and the Robert J. Kleberg, Jr. and Helen C. Kleberg Foundation Medical Research Award.

\newpage
\section{Methods}\label{sec:methods}

\subsection{Choice of Event Speckle Feature}
\label{sec:choice_eSPG_feature}

\begin{figure}[!ht]
    \centering
    \includegraphics[width=0.33\linewidth]{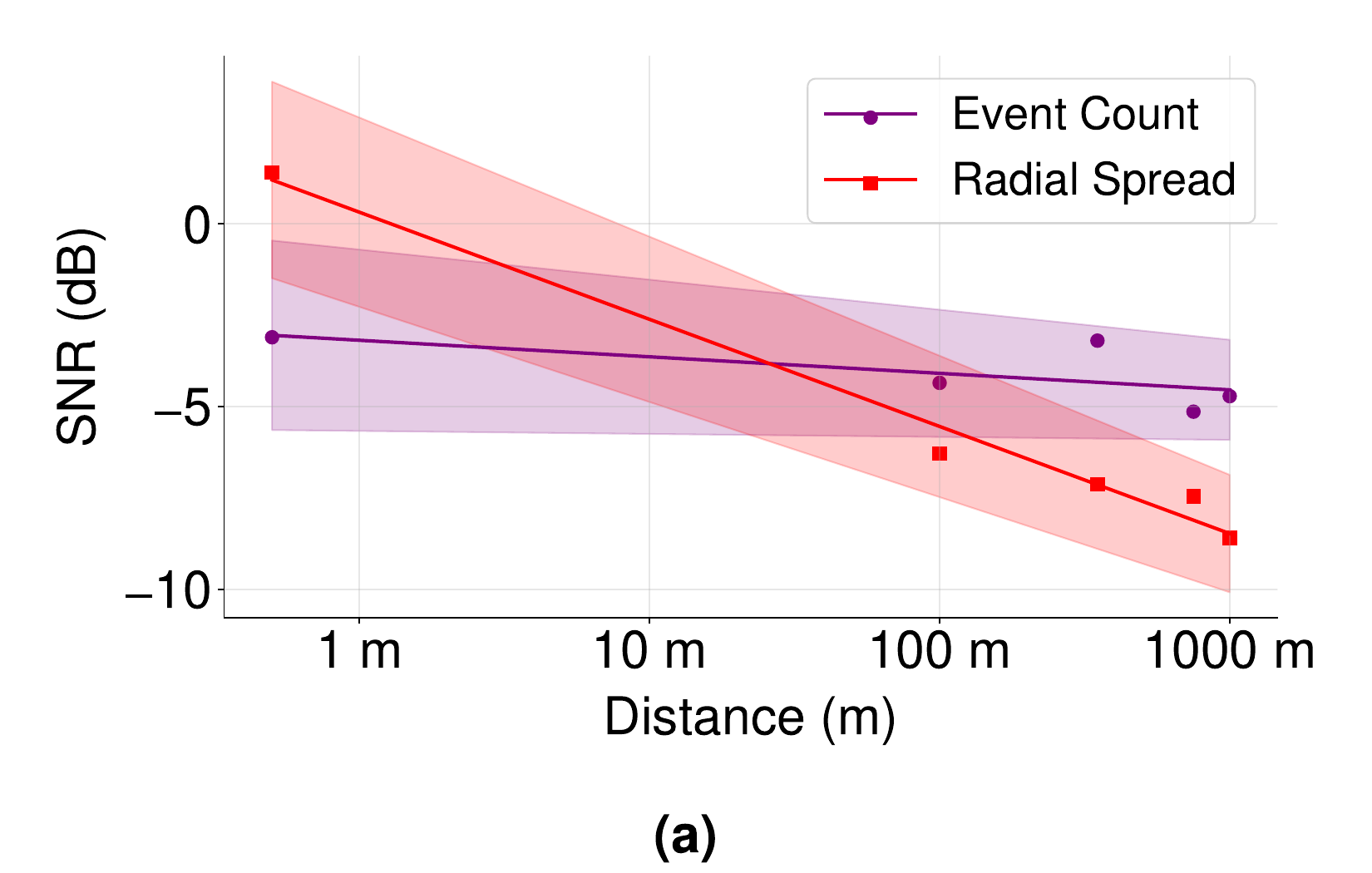}
    \includegraphics[width=0.32\linewidth]{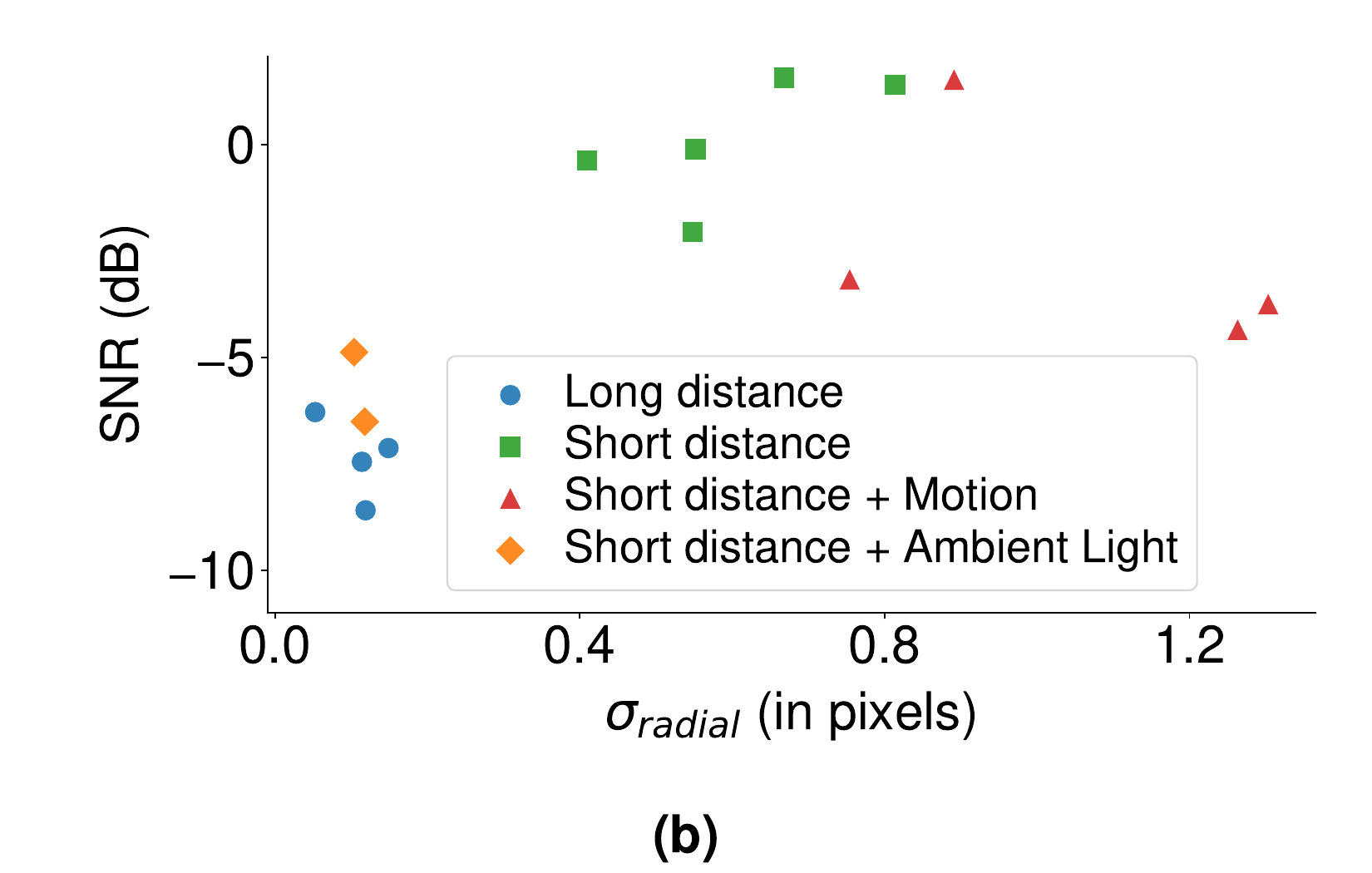}
    \includegraphics[width=0.33\linewidth]{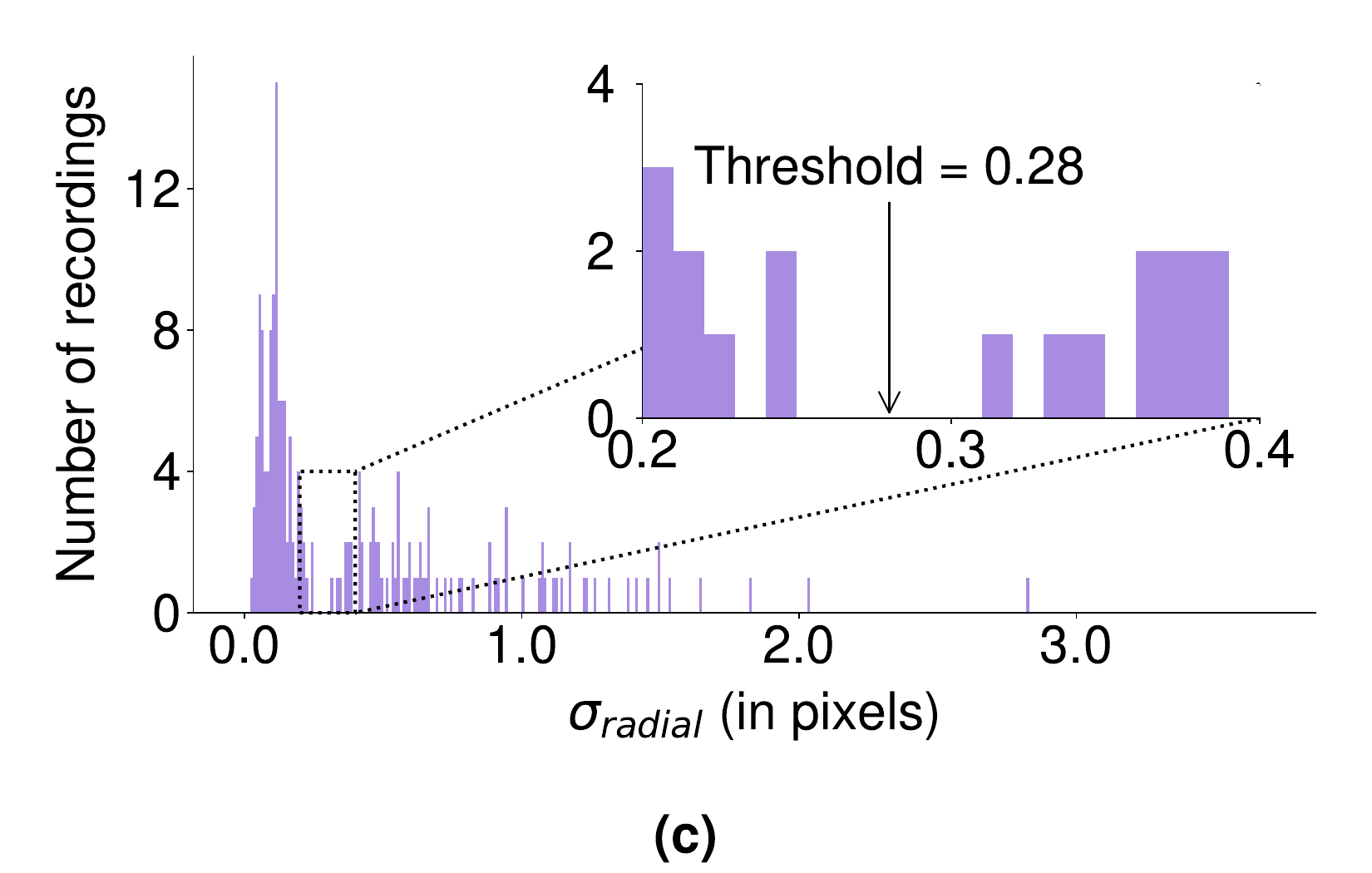}
    \caption{\textbf{\boldmath eSPG can be computed from either event count or radial spread, and the selection can be automated by using a threshold based on the standard deviation of the radial spread of the filtered waveform ($\sigma_{radial}$).}
    \textbf{a.} As the imaging distance increases, the SNR of the waveform obtained from the radial spread feature decreases. \textbf{b.} $\sigma_{radial}$ can be used as a ground-truth-free indicator of the SNR of the radial spread. This measure $\sigma_{radial}$ is the standard deviation of the radial spread of the filtered waveform. As seen in the plot, low $\sigma_{radial}$ values are linked to low SNR and vice versa.
    \textbf{c.} A clear bifurcation can be observed in the histogram of $\sigma_{radial}$, plotted across all trials. We choose $0.28$ as the threshold for our dataset. We use the waveform from the radial feature only if its $\sigma_{radial}$ crosses this threshold.}
    \label{fig:espg_choose}
\end{figure}

\par As seen in \cref{sec:experiments}, the eSPG signal can be computed from either the event count or the radial spread of the events. From \cref{fig:espg_choose}a, we see that the SNR of the radial features drops with the distance, while that of event count stays relatively constant. Thus, the eSPG signal can be calculated from the event count at longer distances and from the radial features at short distances. However, we need a measure that is independent of the ground truth to choose between the two features. Heuristically, we observe that the degree of fluctuation in radial spread indicates the quality of the radial signal. Hence, we plot the standard deviation of the radial spread ($\sigma_{radial}$) against the SNR of radial features for each category. As illustrated in \cref{fig:espg_choose}b, higher $\sigma_{radial}$ typically leads to higher SNR of the radial spread. Therefore, we threshold this $\sigma_{radial}$ to classify whether to use the radial features or not. A histogram plot of these $\sigma_{radial}$ values (\cref{fig:espg_choose}c) reveals a clear separation at $\sim 0.28$, which we choose as our threshold. Recordings with $\sigma_{radial} >  0.28$ use radial spread features, while the rest use event count to estimate the eSPG signal.

\begin{align}
    \sigma_{radial} = \ &\sigma(\text{Butterworth(}r^{eff}_k, \text{freq=HR$_{radial}$, bandwidth=$\pm5~BPM$)}\nonumber\\
    &\quad+\text{Butterworth(}r^{eff}_k, \text{freq=$2\times$HR$_{radial}$, bandwidth=$\pm5~BPM$)}),
\end{align}

where $r^{eff}_k$ is the effective radius (radial spread) as defined in \cref{eq:r_effective}, and HR$_{radial}$ is the heart rate estimated from the radial spread waveform. The Butterworth filters are $1^{\mathrm{st}}$-order bandpass filters applied forward and backward, yielding an effective $2^{nd}$-order response. The forward and backward filtering ensure a zero-phase response.

\subsection{Trade-off Between Speckle Size and SNR}
\label{sec:long_range_tradeoff}

\par In this section, we derive theoretical limits that constrain our imaging optics to ensure adequate sampling and signal levels for speckle imaging. The optical configuration of an imaging system determines the size of the images of point sources at the diffraction limit. Speckles can be considered as random point sources of light (constructive interference), and thus the speckle size $\rho_\text{speckle}$ on the image sensor is given by~\cite{boas2010laser} 
\begin{equation}
    \rho_\text{speckle} = 2.44 \lambda (1+M)f/\#,
\label{eq:speckle_1}
\end{equation}
where $\lambda$ is the wavelength of the coherent illumination source, and $M$ and f/\# are respectively the magnification and the f-number of the optical system being used. Based on the Nyquist sampling criterion, to obtain accurate estimates of speckle contrast, it is recommended that the speckle size be at least two times larger than the pixel size~\cite{kirkpatrick2008detrimental, boas2010laser}, i.e. 
% $\rho_\mathrm{speckle} \ge 2\rho_\mathrm{pixel}$, 
\begin{equation}
    \rho_\mathrm{speckle} = 2.44 \lambda (1+M)f/\# \gtrsim 2\rho_{pixel},
\label{eq:speckle_pixel}
\end{equation}
\noindent
\par If the focal length of the system is $f$ and the aperture is given by $D$, the above equation can also be written as:

\begin{equation}
    \frac{2.44 \lambda (1+M)}{2\rho_\mathrm{pixel}} \gtrsim \frac{D}{f}.
    \label{eq:speckle_pixel_D}
\end{equation}

\par As shown in \cite{kirkpatrick2008detrimental,boas2010laser}, the global speckle contrast is almost directly proportional to pixels per speckle ($N_{pps} = \rho_{speckle}/\rho_{pixel}$) until Nyquist sampling $N_{pps}$ = 2, beyond which there is no significant advantage. Thus lower values of $N_{pps}$ reduce the contrast detection performance linearly by some factor and \cref{eq:speckle_pixel_D} is not a strict bound.

\par \Cref{eq:speckle_pixel} shows that, given a fixed pixel size, increasing f/\# (i.e., decreasing the relative aperture size) increases the speckle size, effectively increasing contrast and, hence, the signal quality.

\par On the other hand, increasing the f/\# also reduces the total energy received by the sensor. Let $P$ denote the optical power actually incident on a target patch of area $A$. Assuming spatially uniform illumination, the mean incident irradiance is

\begin{equation}
\label{eq:incident}
E_{\mathrm{incident}}=\frac{P}{A}.
\end{equation}

\par Let $\rho_{\mathrm{skin}}\in[0,1]$ be the effective diffuse reflectance of the illuminated palm at wavelength $\lambda$. Approximating the palm as Lambertian, its outgoing radiance is independent of direction and equals~\cite{boyd1983radiometry}

\begin{equation}
\label{eq:radiance}
L=\frac{\rho_{\mathrm{skin}}E_{\mathrm{incident}}}{\pi}=\frac{\rho_{\mathrm{skin}}P}{\pi A}.
\end{equation}

\par Let $D$ be the aperture diameter, $f$ the focal length, and $u$ the imaging distance. For an on-axis resolved target at $u\gg f$, the mean sensor-plane irradiance is approximated by the image-irradiance relation~\cite{boyd1983radiometry,greivenkamp2004field}:

\begin{equation}
\label{eq:irradiance}
E_{\mathrm{sensor}}\simeq T\frac{\pi L}{4}\left(\frac{D}{f}\right)^2
=T\frac{\rho_{\mathrm{skin}}P}{4A}\left(\frac{D}{f}\right)^2,
\end{equation}

where $T\in[0,1]$ denotes the optical transmission factor including any atmospheric and optical losses.

\par For monochromatic radiation at wavelength $\lambda$, the corresponding photopic illuminance at the sensor plane is given by

\begin{equation}
\label{eq:lux}
E_{v,\mathrm{sensor}} = K_m V(\lambda) E_{\mathrm{sensor}},
\end{equation}

where $K_m=\SI{683}{\lumen\per\watt}$ is the maximum photopic luminous efficacy, and $V(\lambda)$ is the dimensionless CIE photopic luminous-efficiency function. Combining \cref{eq:irradiance,eq:lux} yields

\begin{equation}
\label{eq:luxfinal}
E_{v,\mathrm{sensor}} \simeq
683\,V(\lambda)\, T \frac{\rho_{\mathrm{skin}}P}{4A} \left(\frac{D}{f}\right)^2\quad\text{lux}.
\end{equation}

\par The factor $683V(\lambda)$ converts monochromatic radiometric irradiance in \si{\watt\per\meter\squared} to photopic illuminance in lux.

\par Let $E_{v,\min}>0$ denote the minimum acceptable sensor-plane illuminance. Requiring $E_{v,\mathrm{sensor}} \geq E_{v,\min}$
and rearranging \cref{eq:luxfinal} yields

\begin{equation}
\label{eq:luxbound}
\frac{D}{f}
\geq \beta_{\mathrm{lux}}
\equiv
\sqrt{
\frac{4A E_{v,\min}}
{K_m V(\lambda) T \rho_{\mathrm{skin}} P}
}\text{ },
\end{equation}

thus establishing a lower bound on the relative aperture (D/f). According to the documentation for the \textit{Prophesee EVK Gen 4} event camera~\cite{prophesee2023evk4}, an illumination level of 5 lux guarantees full imaging characteristics, whereas 0.08 lux represents the low-light cutoff for nominal contrast sensitivity. While higher illumination level is desirable, accumulating events over 25 ms can also improve the signal-to-noise ratio of the extracted pulse waveform, potentially enabling reliable measurements under lower illumination levels. As a first-order estimate, we calculate the sensor-plane illuminance for our 1 km imaging configuration in Supplementary Sec. 2.1[iv]. A detailed complementary analysis of the lower bound, based on noise-floor calculations, is also presented in Supplementary Sec. 2.1[v].

\vspace{10pt}
\noindent
\textbf{Magnification requirement.} Having a sufficiently large ROI or region of pixels illuminated with the speckle pattern is also an important factor in achieving good SPG signal fidelity. In previous literature~\cite{boas2010laser}, it has been noted that speckle contrast is best computed in $7\times7$ spatial windows. For redundancy, if we assume that computing speckle contrast over an area of $25\times25$ pixels ($N\times N$, where $N=25$) yields an SPG signal of good quality, then this imposes constraints upon the focal length of the imaging system. Suppose the palm area of a human hand (which is illuminated by the laser) is $l\times l$ in size in terms of physical length units, and $\rho_\text{pixel}$ is the physical size of a camera pixel, then the image size of the palm must cover at least an ROI of $N\times N$ pixels, that is 
\begin{equation}
    \frac{\text{Image size (in physical units)}}{\text{Pixel pitch}} > N 
    \implies \frac{M \cdot l}{\rho_\text{pixel}} > N
\end{equation}
where $M$ is the magnification of the imaging system. $M$ can be approximated as $f/u$ where $f$ is the focal length, and $u$ is the object distance. Thus, for obtaining good speckle signal quality, we obtain a lower bound on the focal length as follows
\begin{equation}
    f > \frac{N\cdot\rho_\text{pixel}\cdot u}{l}
\label{eq:focal-length-constraint}
\end{equation}

\par \Cref{eq:luxbound,eq:focal-length-constraint} thus indicate that if all the conditions permit, with sufficient laser power, with a large focal length $f$ and a very large aperture $D$, and successful imaging through turbulence, one can obtain HR from eSPG waveforms over a large distance. We provide detailed calculations in Supplementary Sections S2.1 and S2.2 that show how we satisfy this constraint in our long-range setup.

\subsection{Algorithm \& Metrics}
\label{sec:algo_metrics}

\par
\begin{figure}[h]
    \centering
    \includegraphics[width=\textwidth]{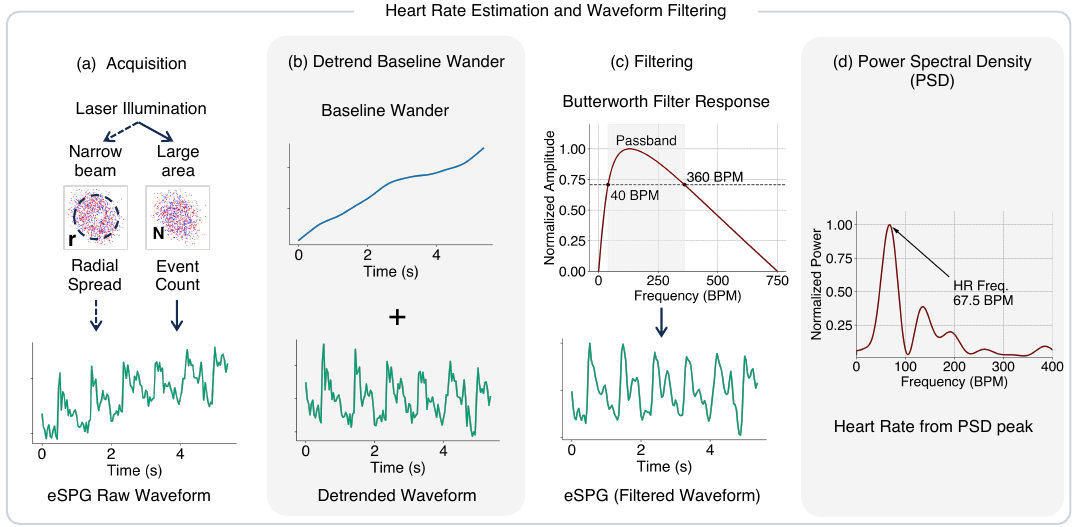}
    \caption{\textbf{Heart rate (HR) can be accurately estimated by detrending and filtering the estimated signals, demonstrated here on the eSPG signal.} \textbf{a,} The raw signal can be extracted from the event stream via statistical features. \textbf{b,} The raw signal is detrended to produce the pulsatile signal and remove the baseline wander. \textbf{c,} The pulsatile signal is denoised with a Butterworth filter to yield the eSPG waveform. \textbf{d,} The filtered waveform encodes the HR, which can be spectrally extracted from its power spectral density.}
    \label{fig:hr_est_proc}
\end{figure}

 Here, we elaborate on the preprocessing steps implemented to clean the signals for HR estimation. The plethysmography waveforms estimated from remote modalities, such as event and frame-based cameras, are inherently noisy. While cases such as ``still'' are less prone to corruption, the other samples are easily corrupted by external factors such as motion, illumination, and distance, as evidenced by \cref{fig:experimental_setup_long,fig:experimental_setup_short_mid}. A summary of our HR estimation workflow is presented in \cref{fig:hr_est_proc}. Before processing these estimated waveforms, we clip the first and last $5\%$ of each raw signal. This is to prevent any edge artifacts from interfering with the steps below. \\

\noindent \textbf{Processing the Estimated Signal.} All plethysmography signals, contact (PPG/SPG) or remote (rPPG/rSPG), are accompanied by baseline wander. Baseline wander here refers to a low-frequency upward or downward shift/trend in the signal caused by factors such as respiration, illumination, body motion, and even subcutaneous vascular changes. In remote sensing, baseline wander is dominated by interfering noise sources, thereby corrupting the signal. Hence, we detrend the signal to improve signal quality. Mathematically, the underlying trend ($x^*$) can be estimated from the raw signal ($y$) via an optimization step:

\begin{align}
x^* = \arg\min_{x \in \mathbb{R}^n} \; \|y - x\|_2^2 \;+\; \lambda^2 \, \|D x\|_2^2.
\end{align}

\par where $D$ is the central second-order difference matrix. Given the differentiability of the above equation, we can derive closed-form solutions for both the trend and the detrended signal ($y^*$). 

\begin{align}
x^* &= \left( I + \lambda^2 D^\top D \right)^{-1}y,\\
y^* &= y - x^*.
\end{align}

\par Following detrending, the signal is bandpass filtered ($40-360~BPM$) to suppress out-of-band noise, using a $1^{\mathrm{st}}$-order Butterworth filter applied forward and backward (i.e., zero-phase filtering), giving an effective $2^{nd}$-order response. The HR is then estimated as the frequency within $40-180~BPM$ that maximizes the sum of the spectral power of the fundamental and its second harmonic. \\

% \subsection{HR Metrics}\label{subsec:hr_metrics}
\noindent \textbf{HR Metrics.} We follow Vilesov et al.~\cite{vilesov2022blending} to quantify the error between the predicted HR ($\hat{h}$) and the ground truth HR ($h$). The definitions for mean absolute error (MAE), root mean squared error (RMSE), mean absolute percentage error (MAPE), and Pearson correlation coefficient ($r$) are as follows:

\begin{align}
    \text{MAE} &= \frac{\sum_{i=1}^{N_{samples}} \left| \hat{h}_i - h_i \right|}{N_{samples}},\\
    \text{RMSE} &= \sqrt{\frac{\sum_{i=1}^{N_{samples}} ( \hat{h}_i - h_i )^{2}}{N_{samples}}} ,\\
    \text{MAPE} &= \frac{1}{N_{samples}}\sum_{i=1}^{N_{samples}} \frac{\lvert \hat{h}_i - h_i \rvert}{h_{i}},\\
    r &= \frac{\sum_{i=1}^{N_{samples}} (\hat{h}_i - \mu_{\hat{h}})(h_i - \mu_z)}
              {\sigma_{\hat{h}}\,\sigma_h\,N_{samples}}.
\end{align}

\par Further, we report the standard errors (SE) for each metric. Let $e_i = \hat{h}_i - h_i$, and $\sigma(\cdot)$ denote the standard deviation.
\begin{align}
    \text{MAE SE}  &= \frac{\sigma\left(|e_i|\right)}{\sqrt{N_{samples}}},\\
    \text{RMSE SE} &= \sqrt{\frac{\sigma\left(e_i^{2}\right)}{\sqrt{N_{samples}}}},\\
    \text{MAPE SE} &= \frac{\sigma\left(|e_i|/h_i\right)}{\sqrt{N_{samples}}},\\
    r\text{ SE}    &= \sqrt{\frac{1 - r^{2}}{N_{samples} - 2}}.
\end{align}

\par Additionally, we report clinical accuracy following the Association for the Advancement of Medical Instrumentation (AAMI) standard~\cite{association2002cardiac}, defined in terms of the sample absolute error ($\text{AE}_i$) and
absolute percentage error ($\text{APE}_i$):
\begin{equation}
    \text{Clinical} = \frac{1}{N_{samples}}\sum_{i=1}^{N_{samples}}
    \mathds{1}\!\left[\, \text{AE}_i \le 5 \;\lor\; \text{APE}_i \le 10\% \,\right].
\end{equation}

\noindent \textbf{Chosen Hyperparameters.} We choose the hyperparameters for our algorithm as shown in \cref{tab:hyperparams}. The camera parameters for data acquisition are shown in \cref{tab:camera_params}.

\begin{table}[!ht]
\centering
\caption{\textbf{Camera specifications and hyperparameters used to estimate plethysmography waveforms and compute the HR from the event- and frame-based cameras.} These configurations are kept constant across all trials and distances. All preprocessing, except the filter order, is kept the same across modalities when computing the HR from the power spectral density.}
\label{tab:hyperparams}
\begin{tabular}{llc}
\toprule
 & Hyperparameter & Value \\
\midrule
\multirow{5}{*}{Common}
 & Detrend                  & True \\
 & Spectral estimator       & Welch \\
 & Pulse-rate band          & $40$--$360~BPM$ \\
 & HR estimate & $\arg\max_{f}\, P(f) + P(2f)$ \\
 & Filtering                & Zero-phase (forward--backward) \\
\midrule
\multirow{6}{*}{Event Camera}
 & Resolution (H$\times$W)      & $720\times1280$ \\
 & Effective sampling (binning) rate & $25~Hz$ \\
 & Event polarity used          & Both \\
 & Radial zones for radial spread   & $10$ \\
 & Butterworth order            & $1$ \\
 & Effective Butterworth order  & $2$ \\
\midrule
\multirow{6}{*}{Frame Camera}
 & Resolution (H$\times$W)      & $1024\times1024$ \\
 & Sampling rate                & $25~Hz$ \\
 & Block size for distPPG and iSPG++                   & $7$~pixels \\
 & Goodness threshold for distPPG and iSPG++  & $0.25$ \\
 & Butterworth order            & $1$ \\
 & Effective Butterworth order  & $2$ \\
\midrule
\multirow{3}{*}{Pulse Oximeter}
 & Sampling rate                & $60$~Hz \\
 & Butterworth order            & $6$ \\
 & Effective Butterworth order  & $12$ \\
\bottomrule
\end{tabular}
\end{table}

\begin{table}[!ht]
\centering
\caption{\textbf{Camera parameters used during acquisition.} Event camera bias parameters: OFF and ON are the parameters that control the contrast thresholds (\texttt{bias\_diff\_off} and \texttt{bias\_diff\_on}). That is, they control the threshold that a pixel's brightness change must exceed to fire an event with negative and positive polarity, respectively. Higher values reduce sensitivity and noise. HPF is the high-pass filter parameter (\texttt{bias\_hpf}); higher values suppress slower changes in brightness. All other biases were set to $0$.}
\label{tab:camera_params}
\begin{tabular}{llcccccc}
\toprule
 & & \multicolumn{4}{c}{Event Camera} & \multicolumn{2}{c}{Frame Camera} \\
\cmidrule(lr){3-6} \cmidrule(lr){7-8}
 & Condition & $f/\#$ & OFF & ON & HPF & $f/\#$ & Exposure ($\mu$s) \\
\midrule
\multirow{4}{*}{Short range}
 & Laser profile (Uniform/Gauss)  & $6$ & $150$ & $100$ & $0$   & $6$ & $1000$ \\
 & Handheld sub-$mW$ laser + hand hovering & $6$ & $80$  & $20$  & $100$ & $6$ & $1000$ \\
 & Ambient light        & $6$ & $80$  & $20$  & $0$   & $6$ & $500$ \\
 & All others        & $6$ & $80$  & $20$  & $0$   & $6$ & $1000$ \\
\midrule
Mid range
 & All            & $1.8$ & $30$ & $-10$ & $0$ & $1.8$ & $1400$ \\
\midrule
\multirow{3}{*}{Long range}
 & $350$~m        & 5 & $30$ & $-10$ & $0$ & 5 & $2000$ \\
 & $750$~m        & 5 & $20$ & $-20$ & $0$ & 5 & $2000$ \\
 & $1$~km         & 5 & $10$ & $-30$ & $0$ & 5 & $2000$ \\
\bottomrule
\end{tabular}
\end{table}

% \bibliography{references}

\clearpage
\begin{center}
  {\LARGE\bfseries Supplementary}
\end{center}
\flushbottom
\maketitle

\thispagestyle{empty}

\beginsupplement

% this line ensures supp sections appear in the table of contents
\addtocontents{toc}{\protect\setcounter{tocdepth}{2}}

\tableofcontents

% \linenumbers

% ============================================================
\newpage
%%%%%%%%%%%%%%%%%%%%%%%%%%%%%%%%%%%%%%%%
%%%%%%%%%%%%%%%%%%%%%%%%%%%%%%%%%%%%%%%%
%%%%%%%%%%%%%%%%%%%%%%%%%%%%%%%%%%%%%%%%

\section{Background and Related Literature}
\label{sec:related_work}

\subsection{Remote Plethysmography (rPPG)}\label{sec:rPPG}

\par Cameras have proven to be effective sensors in sensing and analyzing subtle color changes caused by the pulsation of the blood through the body~\cite{verkruysse2008remote}. Cameras have been widely used to calculate vital signs such as heart rate (HR) and heart rate variability (HRV) estimation via rPPG signals~\cite{de2013robust,wang2016algorithmic,pai2021hrvcam}. One of the early works~\cite{kumar2015distanceppg} used signal-quality metrics within spatial windows to improve signal fidelity from monochrome cameras. This was soon followed by sparse spectral sensing~\cite{magdalena2018sparseppg,nowara2020near}, light transport models~\cite{de2013robust,wang2016algorithmic}, convolutional neural networks (CNNs)~\cite{chen2018deepphys,yu2019remote}, transformers~\cite{yu2022physformer,yu2023physformer++}, state space models~\cite{luo2024physmamba,zou2024rhythmmamba}, contrastive networks~\cite{sun2022contrast,sun2024contrast}, and implicit representations~\cite{chari2024implicit}.

\par Improvements in remote vital sensing with cameras have been further propelled by the addition of other contactless sensing modalities. Vilesov~\etal~\cite{vilesov2022blending} employed late fusion to combine the signals from an RGB camera and frequency-modulated continuous-wave (FMCW) radar to alleviate the effect of melanin on rPPG fidelity~\cite{vilesov2022blending}. Some works~\cite{zhang2016multimodal, joshi2024ibvp} proposed combining RGB and thermal cameras to improve signal fidelity. Additionally, Harish~\etal~\cite{harish2025cogphys} have now shown that HR and HRV estimates from camera-based sensing can be combined with other remote vital-signs data to yield high-level measures such as cognitive load, laying the foundation for future research into more advanced downstream vital-signs sensing.

\par While improvements in rPPG sensing systems have been steady, most rPPG works attempt to sense vitals with cameras placed at close distances, i.e., 3 meters or less~\cite{blackford2016long}. Long-range rPPG was first attempted at distances of 25, 50, 100 meters~\cite{blackford2016long} with heart rate estimation errors steadily increasing from 2 to 10 beats per minute (BPM). Another follow-up study, which involved imaging outdoors under ambient sunlight, tested distances up to 150~m, with errors of 6-8 BPM observed~\cite{blackford2017measurements}.

\subsection{Speckle Plethysmography (SPG)}

Speckle plethysmography (SPG), first coined by Ghijsen \etal~\cite{ghijsen2018wearable}, measures the speckle fluctuations over time to generate a time-varying signal that captures changes in blood flow within cardiac cycles. Several works~\cite{ghijsen2018wearable, dunn2018comparison, herranz2023comparison, schnitzler2024wearable} have analyzed the similarities and differences between PPG and SPG waveforms. Such works have demonstrated that SPG signals are more robust and have a higher signal-to-noise ratio (SNR) than PPG signals. Contact-based SPG devices have also been recently explored for respiratory rate estimation~\cite{hoogeveen2025three} and blood pressure estimation~\cite{lorato2024towards}.

\noindent\textbf{Remote SPG works.} Contactless SPG estimation using a camera and laser placed at a distance was attempted in more recent works~\cite{olazabal2021beat, herranz2022camera, herranz2023comparing, herranz2023multispectral, LeeRollingShutter24}.  Olazabal \etal~\cite{olazabal2021beat} showed that rSPG signals can obtain equally good beat-to-beat intervals as those obtained from contact-based PPG, while being more robust to ambient light changes. Follow-up work~\cite{herranz2022camera} systematically compared rPPG and rSPG in terms of signal quality when used in reflective and transmissive modes, and concluded that transmissive-mode rSPG has `rPPG signal contamination' while reflective-mode rSPG signal quality is better than that of rPPG. Olazabal~\etal~\cite{herranz2023comparing} demonstrated that rSPG-derived pulse wave morphology resembles finger arterial pressure signals more closely than contact-based PPG signals, highlighting the potential of rSPG for non-invasive blood pressure monitoring.

\subsection{Speckle Vibrometry}

Imaging speckle patterns in a defocused setting to measure vibrometry and heartbeats has also been explored~\cite{zalevsky2009simultaneous}. In such methods, the camera is defocused, i.e., it is made to focus in front of or behind the illumination surface. This enables small, tilting-based motions of the laser-illuminated surface to be captured as lateral shifts in the recorded defocused speckle pattern~\cite{zalevsky2009simultaneous}. Any sound or heartbeat, or motion that causes mechanical vibration, is thus captured in the form of displacements of the speckle pattern image. Estimation of heartbeats in a contactless remote setting using speckle vibrometry was first demonstrated in~\cite{zalevsky2009simultaneous}, at distances of up to 60 meters. Several follow-up works demonstrated further capabilities for estimating physiological signals, such as blood pressure~\cite{ozana2015demonstration}, respiration rate~\cite{ozana2015demonstration, havakuk2021validation, que2024speckle}, and pulse wave velocity~\cite{sirkis2016monitoring}. More recently, speckle vibrometry has demonstrated detection of cardiac motions through cm-thick textiles~\cite{que2022contactless}, and has been employed in a multimodal setup alongside radar sensing for patient monitoring~\cite{zhu2024camera}. Such defocused speckle sensing, or speckle vibrometry, relies on accurate measurement of speckle pattern displacements through optical flow, and inherently senses mechanical displacements or vibrations. In this work, our proposed methods do not rely on defocusing the camera, and we aim to exploit the dynamic \textit{speckle pattern change} rather than \textit{speckle pattern displacements}.

\subsection{Event-based Imaging}

Frame-based intensity cameras suffer from low frame rate, low dynamic range, and higher memory and power consumption.
Event cameras~\cite{gallego2020event,lichtsteiner2008128,brandli2014240,posch2010qvga} are biologically inspired neuromorphic sensors that have emerged as exciting alternatives to frame cameras.
They asynchronously measure per-pixel changes in log intensity (known as events) when they exceed a predefined threshold.
Their high temporal resolution ($\mu s$-scale latency) and high dynamic range ($>$ 120 dB) make them particularly well-suited for capturing fast, low-amplitude intensity variations.
These properties have enabled a wide range of applications \cite{gallego2020event,chakravarthi2024recent}, including optical flow estimation~\cite{zhu2019unsupervised,gallego2018unifying}, robotics~\cite{mueggler2017event,falanga2020dynamic} autonomous driving~\cite{maqueda2018event,gehrig2024low}, 3D reconstruction \cite{matsuda2015mc3d,muglikar2021esl,dashpute2026accurate}, and microscopy \cite{guo2024eventlfm,cabriel2023event}.

\noindent
\textbf{Event-based Speckle Imaging.}
Event cameras have recently been used in laser speckle imaging, primarily exploring two directions: laser vibrometry, where surface vibrations are recovered from the movement of speckle \cite{cai2025event2audio,zhou2025event}, and biological flow imaging, where speckle decorrelation is used to probe motion in scattering media \cite{ge2020dynamic,balaji2024probing}.

Recent work \cite{jagtap2023heart} uses physical marking on the skin on the wrist (radial artery) to detect micro-motions of the artery and detect heart rate. Concurrent work~\cite{feng2026empd} uses laser illumination to explore event-based plethysmography on the wrist (radial artery) using a learning-based approach, but it does not explicitly model speckle formation or its underlying physical dynamics, limiting interpretability and generalizability. These approaches are typically limited to controlled or short-range settings and do not explicitly leverage the temporal contrast structure of speckle to robustly extract physiological signals. Using our proposed eSPG method, we obtain high-fidelity SPG waveforms to estimate not only heart rate (HR) but also heart rate variability (HRV) in our short-range imaging experiments.  

%%%%%%%%%%%%%%%%%%%%%%%%%%%%%%%%%%%%%%%%
%%%%%%%%%%%%%%%%%%%%%%%%%%%%%%%%%%%%%%%%
%%%%%%%%%%%%%%%%%%%%%%%%%%%%%%%%%%%%%%%%

\newpage
\section{Experimental Setup}

\subsection{Satisfying Speckle size-SNR tradeoff}

Speckle size $\rho_{speckle}$ \cite{boas2010laser} is given by 

\begin{equation}
    \rho_{speckle} = 2.44 \lambda (1+M)f/\#,
\end{equation}

where $\lambda$ is the wavelength of the coherent illumination source, and $M$ and $f/\#$ are respectively the magnification and the $f$-number of the optical system being used.

Based on the Nyquist sampling criterion, it is recommended that the speckle size be two times larger than the pixel size, i.e., $\rho_{speckle} \gtrsim 2\rho_{pixel}$. Failure to meet this criterion causes the effective contrast measurement to drop linearly, which may sometimes be necessary in low-light regimes, where a higher relative aperture size is required.

\vspace{10pt}
\noindent
\textbf{\boldmath [i] $0.6~m~(<1~m)$ imaging distance.}\\

\noindent
For our experimental setup for imaging at short distance ($60~cm$) with NIR laser $\lambda = 0.850~\mu m$, focal length $f = 12.5~mm$, magnification $M \approx 0.02$, the upper bound on $D/f$ (right-hand side constraint on $D/f$ from Eq. (22) of the main paper) is:

\begin{equation}
\frac{2.44 \lambda (1+M)}{2\rho_\mathrm{pixel}} = \frac{2.44 \times 0.850 \times (1+0.02)}{2 \times 4.86} = 0.218,
\end{equation}

where our $D/f$ ($=1/f/\#$) is $1/6 = 0.167 < 0.218$. This satisfies the right-hand side constraint of Eq. (22) of the main paper.

\vspace{10pt}
\noindent
\textbf{\boldmath [ii] $100~m$ imaging distance.}\\

\noindent
For our setup for mid-range imaging at 100 m with red laser $\lambda = 0.633~\mu m$, focal length $f$ = 135 mm, f/\# = f/1.8, magnification $M \approx 0.00135$, the upper bound on $D/f$ is:

\begin{equation}
\frac{2.44 \lambda (1+M)}{2\rho_\mathrm{pixel}} = \frac{2.44 \times 0.633 \times (1+0.00135)}{2 \times 4.86} = 0.159,
\end{equation}

\noindent
where our D/f (=1/f/\#) is 1/1.8 = 0.55 $\nless$ 0.159.
In other words, the speckle size is
\begin{equation}
    \rho_\text{speckle} = 2.44 \lambda (1+M)f/\# = 2.44 \times 0.633 \times (1+0.00135) \times 1.8 = 2.78~\mu m < 2 \times \rho_{pixel} = 9.72~\mu m
\end{equation}
This means that we suffer speckle contrast by a factor as mentioned in Boas et al. \cite{boas2010laser}. This is done to maximize the signal-to-noise ratio in this low-light regime.

\vspace{10pt}
\noindent
\textbf{\boldmath [iii] $1~km$ imaging distance.}\\

\noindent
For long-range imaging setup for 1 km with red laser $\lambda = 0.633~\mu m$, focal length $f$ = 1500 mm, f/\# = f/5, magnification M $\approx$ 0.0015, the upper bound on D/f is:

\begin{equation}
    \frac{2.44 \lambda (1+M)}{2\rho_\mathrm{pixel}} = \frac{2.44 \times 0.633 \times (1+0.0015)}{2 \times 4.86} = 0.159,
\end{equation}

\noindent
where our D/f (=1/f/\#) is 1/5 = 0.2 $\nless$ 0.159. The speckle size is:

\begin{equation}
\rho_\text{speckle} = 2.44 \lambda (1+M)f/\# = 2.44 \times 0.633 \times (1+0.0015) \times 5 = 7.73~\mu m < 2 \times \rho_{pixel} = 9.72~\mu m,    
\end{equation}

\newpage
\noindent
\textbf{\boldmath [iv] Illuminance-based lower bound for $1~km$ imaging distance.}\\

\vspace{10pt}
\noindent
\textbf{Parameter values and assumptions.} Table~\ref{tab:parameters1} lists the parameters used for the numerical example.
%--------------------------------------------------------------------------------
\begin{table}[htbp]\centering\small
\caption{Inputs for the 1 km sensor-plane illuminance estimate.}\label{tab:parameters1}
\begin{tabularx}{\linewidth}{@{}l l X@{}}\toprule
Symbol & Value & Definition and qualification \\\midrule
$u$ & \SI{1000}{\meter} & Imaging distance. \\
$f$ & \SI{1.5}{\meter} & Telescope focal length. \\
$f/D$ & 5 & Telescope f-number. \\
$D$ & \SI{0.300}{\meter} & Entrance pupil diameter, $D=f/5$. \\
$P$ & \SI{5}{\milli\watt} & Laser power; assumed to fully reach the palm. \\
$A$ & \SI{0.010}{\meter\squared} & Approximately \SI{10}{\centi\meter}$\times$\SI{10}{\centi\meter} illumination area. \\
$\lambda$ & \SI{633}{\nano\meter} & Monochromatic red illumination. \\
$\rho_{\mathrm{skin}}$ & 0.50 & Assumed effective Lambertian palm reflectance. \\
$T$ & 0.50 & Assumed target-to-sensor optical and atmospheric transmission. \\
$V(\lambda)$ & 0.23 & Approximate CIE photopic luminous efficiency at \SI{633}{\nano\meter}; \\
$K_m$ & \SI{683}{\lumen\per\watt} & Standard photopic luminous-efficacy constant. \\
\bottomrule\end{tabularx}
\end{table}
%--------------------------------------------------------------------------------

\vspace{10pt}
\noindent
\textbf{Numerical substitution and verification.}
From the telescope f-number,
\begin{align}
D&=\frac{\SI{1.5}{\meter}}{5}=\SI{0.300}{\meter},\\
\frac{D}{f}&=\frac{0.300}{1.5}=0.200.
\end{align}
The mean irradiance on the illuminated palm is
\begin{equation}
E_{\mathrm{incident}}=\frac{\SI{0.005}{\watt}}{\SI{0.010}{\meter\squared}}=\SI{0.500}{\watt\per\meter\squared}.
\end{equation}
The corresponding Lambertian radiance is
\begin{equation}
L=\frac{(0.50)(0.500)}{\pi}\simeq\SI{7.958e-2}{\watt\per\meter\squared\per\steradian}.
\end{equation}
For $T=0.5$, the sensor-plane irradiance is
\begin{align}
E_{\mathrm{sensor}}
&= \frac{(0.50)(0.50)(\SI{0.005}{\watt})}{4(\SI{0.010}{\meter\squared})}\left(\frac{0.300}{1.5}\right)^2\notag\\
&= \SI{1.250e-3}{\watt\per\meter\squared}.
\end{align}
Using $V(\SI{633}{\nano\meter})\simeq0.23$,
\begin{align}
E_{v,\mathrm{sensor}}
&= 683(0.23)(1.250\times10^{-3})\notag\\
&= \boxed{\SI{0.19636}{\lux}}.
\label{eq:numericlux}
\end{align}
For the idealized transmission $T=1$, the sensor-plane irradiance and illuminance become
\begin{align}
E_{\mathrm{sensor}}\big|_{T=1}&=\SI{2.500e-3}{\watt\per\meter\squared},\\
E_{v,\mathrm{sensor}}\big|_{T=1}&=683(0.23)(2.500\times10^{-3})\notag\\
&=\boxed{\SI{0.39273}{\lux}}.
\end{align}

\noindent
\textit{Prophesee EVK4} reports a non-guaranteed low-light cutoff of 0.08 lux, described as the minimum light level guaranteeing nominal contrast sensitivity, and a 5 lux condition guaranteeing inclusion of all possible operating points \cite{prophesee2023evk4}.
Consequently, comparing the calculated 0.196-0.392 lux \emph{sensor-plane photopic equivalent} with the quoted \SI{0.08}{\lux} cutoff validates the minimum lighting condition performance. Our performance and the quality of the pulse waveform are further improved by event accumulation over 25 ms. In our experiments, we cap the laser power at $\lesssim$ 5~mW, but the per-unit-area illumination is well below the laser safety limit and can be increased further to improve performance.

% \newpage
\vspace{10pt}
\noindent
\textbf{\boldmath [v] Noise floor calculations for lower bound on $D/f$}\\

Let $P$ (in \si{\watt}) denote the optical power actually incident on a target patch of area $A$ (\si{\meter\squared}). In our case, P is the laser power and A is the illumination area on the palm. Assuming illumination is approximately spatially uniform, the mean incident irradiance $E_{\mathrm{incident}}$ (\si{\watt\per\meter\squared}) is
\begin{equation}
E_{\mathrm{incident}}=\frac{P}{A}.
\label{eq:incident_supp}
\end{equation}
Let $\rho_{skin} \in [0,1]$ be the effective diffuse reflectance of the illuminated palm at optical wavelength $\lambda$ (\si{\meter}). Approximating the palm as a Lambertian reflector, its outgoing radiance $L$ (\si{\watt\per\meter\squared\per\steradian}) is independent of observation direction and equals \cite{boyd1983radiometry}:
\begin{equation}
L=\frac{\rho E_{\mathrm{inc}}}{\pi}=\frac{\rho_{skin} P}{\pi A}.
\label{eq:radiance_supp}
\end{equation}
The skin is not truly Lambertian; $\rho_{skin}$ is therefore a radiometric approximation.

Let $D$ (\si{\meter}) be the aperture diameter, $f$ (\si{\meter}) its focal length, and $u$ (\si{\meter}) the imaging distance.
For an on-axis, resolved target, $u\gg f$, the sensor-plane irradiance $E_{\mathrm{sensor}}$ (\si{\watt\per\meter\squared}) follows the standard image-irradiance relation \cite{boyd1983radiometry,greivenkamp2004field}, while also being affected by optical transmission through atmospheric propagation (assuming a factor $T$):
\begin{equation}
\label{eq:irradiance_supp}
E_{\mathrm{sensor}} \simeq T \frac{\pi L}{4}\left(\frac{D}{f}\right)^2
= T \frac{\rho_{skin} P}{4A}\left(\frac{D}{f}\right)^2.
\end{equation}
Let $\ell$ (\si{\meter}) denote the pitch of a square sensor pixel. Assuming that its photoreceptive area equals $\ell^2$, the mean incident optical power per pixel, $P_{\mathrm{px}}$ (\si{\watt}) is
\begin{equation}
\label{eq:pixelpower}
P_{\mathrm{px}} = E_{\mathrm{sensor}} \ell^2.
\end{equation}
Let $\tau$ (\si{\second}) be an approximate photon-counting observation interval; this is explicitly defined for frame cameras but not for event cameras and is approximated later in calculations. The optical energy $U_{\mathrm{px}}$ (\si{\joule}) incident during this interval is
\begin{equation}
\label{eq:energy}
U_{\mathrm{px}} = P_{\mathrm{px}}\tau = E_{\mathrm{sensor}} \ell^2 \tau.
\end{equation}
Each photon at wavelength $\lambda$ has energy $\varepsilon_\gamma=hc/\lambda$, where $h$ (\si{\joule\second}) is Planck's constant and $c$ (\si{\meter\per\second}) is the speed of light in vacuum. The mean number of incident photons $N_\gamma$ is therefore
\begin{equation}
\label{eq:photons}
N_\gamma = \frac{U_{\mathrm{px}}}{hc/\lambda}
= \frac{E_{\mathrm{sensor}}\ell^2\tau\lambda}{hc}.
\end{equation}
Let $\eta \in [0,1]$ denote the sensor quantum efficiency at $\lambda$. With no additional modeled losses, the mean detected photoelectron count $N_e$ is
\begin{align}
\label{eq:electrons}
N_e
&=\eta N_\gamma = \frac{\eta E_{\mathrm{sensor}}\ell^2\tau\lambda}{hc} \notag\\
&=\frac{T\eta\rho_{skin} P\lambda\tau}{4Ahc}
\left(\ell\frac{D}{f}\right)^2.
\end{align}

Now, let $C\geq0$ denote the fraction of optical-intensity modulation relative to its mean. In a simplified model, the modulation amplitude $\Delta N_e$ (photoelectrons) is
\begin{equation}\label{eq:modulation}
\Delta N_e = CN_e.
\end{equation}
This definition assumes average light modulation but doesn't explicitly account for speckle dynamics.

For independent Poisson photon arrivals,
the standard deviation of photon shot noise, $\sigma_{\mathrm{shot}}$ (photoelectrons rms), is
\begin{equation}
\label{eq:shot}
\sigma_{\mathrm{shot}}=\sqrt{N_e}.
\end{equation}
Assuming shot noise as the dominant source of noise and calculating SNR, we get:
\begin{equation}
\label{eq:snr}
\mathrm{SNR}
\equiv \frac{\Delta N_e}{\sigma_{\mathrm{shot}}}
= \frac{CN_e}{\sqrt{N_e}}
= C\sqrt{N_e}.
\end{equation}
Let $\mathrm{SNR}_{\min}>0$ be the chosen minimum acceptable dimensionless modulation SNR. Requiring $\mathrm{SNR} \geq \mathrm{SNR}_{\min}$ and substituting \cref{eq:electrons} gives
\begin{equation}
\label{eq:requirement}
C\sqrt{\frac{\eta T \rho_{skin} P\lambda\tau}{4Ahc}
\left(\ell\frac{D}{f}\right)^2}\geq \mathrm{SNR}_{\min}.
\end{equation}
Since all relevant parameters are nonnegative, isolating the optical parameter yields
\begin{equation}
\label{eq:bound}
\boxed{
\frac{D}{f}
\geq \beta_{\mathrm{noise}}
\equiv \frac{\mathrm{SNR}_{\min}}{\ell C} \sqrt{\frac{4Ahc}{\eta T \rho_{skin} P\lambda\tau}}.
}
\end{equation}
Increasing pixel pitch or relative aperture improves the ideal photon budget, while increasing aperture also decreases image-plane speckle size; the latter establishes a separate upper bound as shown in the Methods section of the main paper.

\vspace{5pt}
\noindent
We calculate the approximate bounds for the 1 km imaging distance case below.

\vspace{5pt}
\noindent
\textbf{Parameter values and assumptions.} \Cref{tab:parameters} shows the parameter values and marks the assumptions made in picking up some values like $\rho_{skin}$ and the quantum efficiency of the event camera.
\begin{table}[!h]
\centering\small
\caption{Inputs for the \SI{1}{km} SNR-constraint calculation.}
\label{tab:parameters}
\begin{tabularx}{\linewidth}{@{}l l X@{}}
\toprule
Symbol & Value & Value and physical significance \\
\midrule
$u$ & \SI{1000}{\meter} & Imaging distance.\\
$f$ & \SI{1.5}{\meter} & Telescope focal length.\\
$f/D$ & 5 & f-number; determines collection cone.\\
$D$ & \SI{0.300}{\meter} & Derived from $D=f/5$; entrance-pupil diameter.\\
$P$ & \SI{5}{\milli\watt} & Laser power; assumed to fully reach the palm.\\
$A$ & \SI{0.010}{\meter\squared} & Approximately \SI{10}{\centi\meter}$\times$\SI{10}{\centi\meter} measured spot; uniform illumination approximation.\\
$\lambda$ & \SI{633}{\nano\meter} & Red laser wavelength.\\
$\ell$ & \SI{4.86}{\micro\meter} & EVK4-HD/IMX636 pixel pitch from sensor documentation.\\
$\rho_{skin}$ & 0.50 & \textit{Assumed} effective diffuse palm reflectance~\cite{jablonski2000evolution}.\\
$\eta$ & 0.50 & \textit{Assumed} average QE at \SI{633}{\nano\meter}.\\
$C$ & 0.50 & \textit{Approximate} fractional single-pixel intensity modulation.\\
$\tau$ & \SI{0.010}{\second} & \textit{Assumed} effective counting interval.\\
$\mathrm{SNR}_{\min}$ & 5 & \textit{Chosen} illustrative minimum SNR.\\
$T$ & 1 & \textit{Assumed} ideal transmission of unit optical and atmospheric transmission.\\
\bottomrule
\end{tabularx}
\end{table}

\noindent
\textbf{Numerical substitution and verification.}
Inserting the parameter values from \cref{tab:parameters},
the aperture diameter and relative aperture are
\begin{align}
D &= \frac{\SI{1.5}{\meter}}{5}
   = \SI{0.300}{\meter},\\
\frac{D}{f}
  &= \frac{0.300}{1.5}
   = \boxed{0.200}.
\label{eq:actual}
\end{align}
The sensor irradiance and incident optical power per pixel are
\begin{align}
E_{\mathrm{sensor}}
&= \frac{(1)(0.50)(\SI{0.005}{\watt})}
        {4(\SI{0.010}{\meter\squared})}
   \left(\frac{0.300}{1.5}\right)^2
\notag\\
&= \SI{2.50e-3}{\watt\per\meter\squared},\\
P_{\mathrm{px}}
&= (\SI{2.50e-3}{\watt\per\meter\squared})
   (\SI{4.86e-6}{\meter})^2
\notag\\
&= \SI{5.9049e-14}{\watt}.
\end{align}
The incident photon rate $\dot N_\gamma$ and generated
photoelectron rate $\dot N_e$ are
\begin{align}
\dot N_\gamma
&= \frac{P_{\mathrm{px}}\lambda}{hc}
\notag\\
&= \frac{
(5.9049\times10^{-14})(633\times10^{-9})
}{
(6.62607015\times10^{-34})(299792458)
}
\notag\\
&\simeq \SI{1.88165e5}{\per\second},\\
\dot N_e
&= \eta\dot N_\gamma
= (0.50)(1.88165\times10^5)
\notag\\
&\simeq \SI{9.40826e4}{\per\second},\\
N_e
&= \dot N_e\tau
= (9.40826\times10^4)(0.010)
\notag\\
&\simeq 940.83\ \text{photoelectrons}.
\end{align}
Substituting the assumed quantities into
\cref{eq:bound}, with
$h=\SI{6.62607015e-34}{\joule\second}$ and
$c=\SI{299792458}{\meter\per\second}$, gives
\begin{align}
\beta_{\mathrm{noise}}
&= \frac{\mathrm{SNR}_{\min}}{\ell C}
\sqrt{
\frac{4Ahc}
{\eta T\rho_{skin}P\lambda\tau}
}
\notag\\
&= \frac{5}{(4.86\times10^{-6})(0.50)}
\sqrt{
\frac{
4(0.010)(6.62607015\times10^{-34})(299792458)
}{
(0.50)(1)(0.50)(0.005)
(633\times10^{-9})(0.010)
}
}
\notag\\
&\simeq \boxed{0.06520}.
\end{align}
The final comparison is
\begin{equation}
\label{eq:comparison}
D/f = 0.200
>
\beta_{\mathrm{noise}} = 0.06520.
\end{equation}
Thus, the assumed optical configuration exceeds the
idealized lower bound by a factor of approximately
$0.200/0.06520\simeq3.067$.
\noindent
Equivalently, \cref{eq:snr} gives
\begin{equation}
\mathrm{SNR}_{\mathrm{mod}}
= C\sqrt{N_e}
= 0.50\sqrt{940.83}
\simeq 15.34,
\end{equation}
compared with the chosen threshold
$\mathrm{SNR}_{\min}=5$.
\noindent
For a smaller value of $\tau$=2\si{ms}, and lower optical transmission of T=0.5, the bound is barely satisfied with $\beta_{noise} \approx 0.2$. As described in the above subsection [iv], the laser power can be further increased to improve the performance in noise-limited regimes.

%%%%%%%%%%%%%%%%%%%%%%%%%%%%%%%%%%%%%%%%%%%%%%
%%%%%%%%%%%%%%%%%%%%%%%%%%%%%%%%%%%%%%%%%%%%%%
%%%%%%%%%%%%%%%%%%%%%%%%%%%%%%%%%%%%%%%%%%%%%%

% 

%--------------------------------------------------------------------------
%--------------------------------------------------------------------------

\newpage
\subsection{Satisfying Magnification Constraints}

\vspace{10pt}
As shown in the main paper, based on the magnification requirement, the focal length $f$
\begin{equation}
    f > \frac{N\rho_{pixel} u}{l}
\label{eq:focal-length-constraint_supp}
\end{equation}
Let's say we need at least 25 $\times$ 25 pixels on the camera sensor, making $N=25$. The sensor size is $\rho_{pixel} = 4.86~\mu m$, illumination area is approximately 10~cm $\times$ 10~cm, making $l$ = 10~cm = 0.1~m. Below we show the constraints for different distances $u$ and the focal lengths $f$ of the respective optics.

\vspace{10pt}
\noindent
\textbf{100 m imaging distance.}\\

% \noindent
Here $u$ = 100~m, $f$ = 135~mm = 0.135~m.

% \noindent
So ${N \rho_{pixel} u}/{l} = (25) (4.86 \times 10^{-6}) (100)/(0.1)~m = 0.121~m = 121~mm$, and our chosen focal length $f$ = 135~mm > 121~mm.

\vspace{10pt}
\noindent
\textbf{350 m imaging distance.}\\

% \noindent
Here $u$ = 350~m, $f$ = 1500~mm = 1.5~m.

% \noindent
So ${N \rho_{pixel} u}/{l} = (25) (4.86 \times 10^{-6}) (350)/(0.1)~m = 0.426~m = 426~mm$, and our chosen focal length $f$ = 1500~mm > 426~mm.

\vspace{10pt}
\noindent
\textbf{1000 m imaging distance.}\\

% \noindent
Finally, for 1~km, $u$ = 1000~m, $f$ = 1500~mm = 1.5~m.

% \noindent
So ${N \rho_{pixel} u}/{l} = (25) (4.86 \times 10^{-6}) (1000)/(0.1)~m = 1.215~m = 1215~mm$, and our chosen focal length $f$ = 1500~mm > 1215~mm.

\vspace{10pt}
\noindent
Hence, we satisfy the minimum magnification requirement set based on the minimum of 25 $\times$ 25 required pixel region of interest for the recordings.

%--------------------------------------------------------------------------
%--------------------------------------------------------------------------

\newpage
\subsection{Setups for Ambient Light and Motion Robustness Study}

\begin{figure}[ht]
    \centering
    \includegraphics[width=0.5\linewidth]{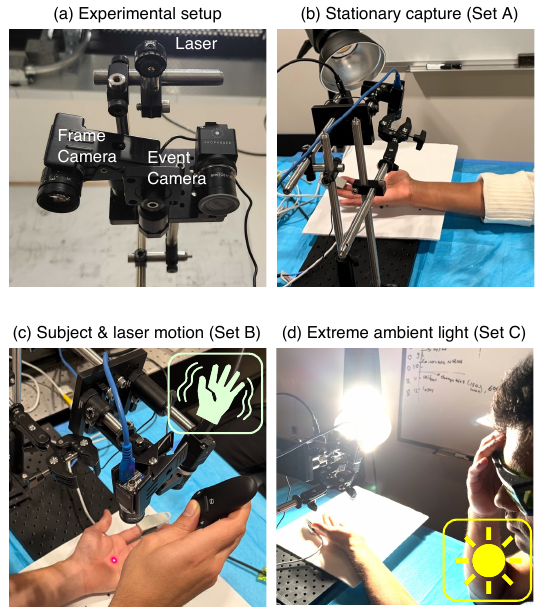}
    \caption{\textbf{Short-range ($<1m$) experiments: Our proposed method is robust to motion and ambient light.} (a) Shows the experimental setup. (b) In Set A, we capture the palm measurement while the palm is still. (c) In Set B, we aim a classroom pointer laser at the palm and hover the palm, simulating a jittering motion. (d) In Set C, we flood-illuminate with large amounts of ambient light to further test robustness. Our method performs best across all the above conditions.}
    \label{fig:short-range-setup}
\end{figure}

\vspace{10pt}
\Cref{fig:short-range-setup}a-d show all the short-distance setups that test the robustness of our system against motion and ambient light.

\newpage
\section{Additional Experiments \& Results}

\subsection{Ablation: eSPG Estimation on the Dorsal Side}\label{subsec:dorsal_experiments}

\par In addition to measuring the speckle signals on the palmar side of the hand, we also capture readings from the dorsal side of the hand for the short-range experiment. We tabulate these readings in \cref{tab:dorsal_hr}. Here, the palmar side recordings are the same as those labeled ``still'' in the main paper. Across the 10 participants, we find that palmar readings have lower error than dorsal readings, owing to higher perfusion on the palmar side~\cite{sulli2014evaluation}. That said, 9/10 dorsal recordings clear the clinical threshold test introduced by AAMI~\cite{association2002cardiac}, with an MAE of $1.7~BPM$. This indicates that while the palmar side is better suited for SPG sensing, the dorsal side remains viable when sensing the palm is not possible.

\begin{table*}[ht]
\centering
\caption{\textbf{eSPG signals can also be estimated from the dorsal side of the palm.} We compare the eSPG signals sensed from the dorsal side against the palmar side (labeled as ``still'' in the main paper)}. 
\begin{tabular}{c l c c c c c}
\toprule
 & & \multicolumn{4}{c}{HR Error ($BPM$)}&\multicolumn{1}{c}{Clinical}\\
\cmidrule(lr){3-6} \cmidrule(lr){7-7}
 & Method & MAE $\downarrow$ & RMSE $\downarrow$ & MAPE(\%) $\downarrow$ & $r$ $\uparrow$ & AAMI $\uparrow$ \\
\midrule
\multirow{4}{*}{\rotatebox{90}{\textbf{Palmar}}} & distPPG & $24.6 \pm 4.9$ & $29.0 \pm 17.7$ & $37.7 \pm 9.0$ & $-0.20 \pm 0.35$ & $1/10$ \\
 & iSPG & \winner{$0.7 \pm 0.5$} & \winner{$1.8 \pm 1.8$} & \winner{$0.9 \pm 0.6$} & \winner{$0.99 \pm 0.05$} & \winner{$10/10$} \\
 & \textbf{iSPG++} (ours) & \runnerup{$3.9 \pm 3.0$} & \runnerup{$10.4 \pm 10.0$} & $5.3 \pm 4.2$ & \runnerup{$0.71 \pm 0.25$} & \runnerup{$9/10$} \\
\cmidrule(lr){2-7}
 & \textbf{eSPG} (ours) & \winner{$0.7 \pm 0.5$} & \winner{$1.8 \pm 1.8$} & \runnerup{$1.0 \pm 0.6$} & \winner{$0.99 \pm 0.05$} & \winner{$10/10$} \\
\midrule
\multirow{4}{*}{\rotatebox{90}{\textbf{Dorsal}}} & distPPG & $22.7 \pm 4.7$ & $27.2 \pm 14.9$ & $33.5 \pm 7.2$ & $0.20 \pm 0.35$ & \runnerup{$3/10$} \\
 & iSPG & \winner{$1.5 \pm 1.1$} & \winner{$3.8 \pm 3.7$} & \winner{$2.0 \pm 1.5$} & \winner{$0.95 \pm 0.11$} & \winner{$9/10$} \\
 & \textbf{iSPG++} (ours) & \runnerup{$1.6 \pm 1.1$} & \runnerup{$3.9 \pm 3.8$} & \runnerup{$2.1 \pm 1.5$} & \winner{$0.95 \pm 0.12$} & \winner{$9/10$} \\
\cmidrule(lr){2-7}
 & \textbf{eSPG} (ours) & \runnerup{$1.6 \pm 1.2$} & $4.0 \pm 3.9$ & \runnerup{$2.1 \pm 1.5$} & \runnerup{$0.94 \pm 0.12$} & \winner{$9/10$} \\
\bottomrule
\end{tabular}
\label{tab:dorsal_hr}
\end{table*}

\newpage
\subsection{Vascular Occlusion}\label{subsec:cuff_experiments}

\par Having established the theoretical basis for capturing hemodynamic changes in blood flow, we validate our claims via a vascular occlusion test. Specifically, we test our imaging setup to distinguish complete arterial occlusion from non-occlusion. For the measurement, the participant wears a cuff. We then capture a $2~min$ recording of the participant's palm while it is kept steady. The recording is divided into 5 phases. The first $30~secs$ constitute the pre-occlusion phase and are recorded as the participant's baseline reading. Following this, the cuff pressure is ramped up from $0$ to $140~mm$ Hg over $15~secs$. This phase lets us achieve complete arterial occlusion. In the third phase, we maintain cuff pressure for $30~secs$, ensuring complete vascular occlusion and the absence of a pulse. In the last two phases, we release the pressure from the cuff over $15~secs$, followed by a $30~secs$ recording under no-occlusion ($0~mm$~Hg).

\begin{figure}[!ht]
    \centering
    \includegraphics[width=0.8\linewidth]{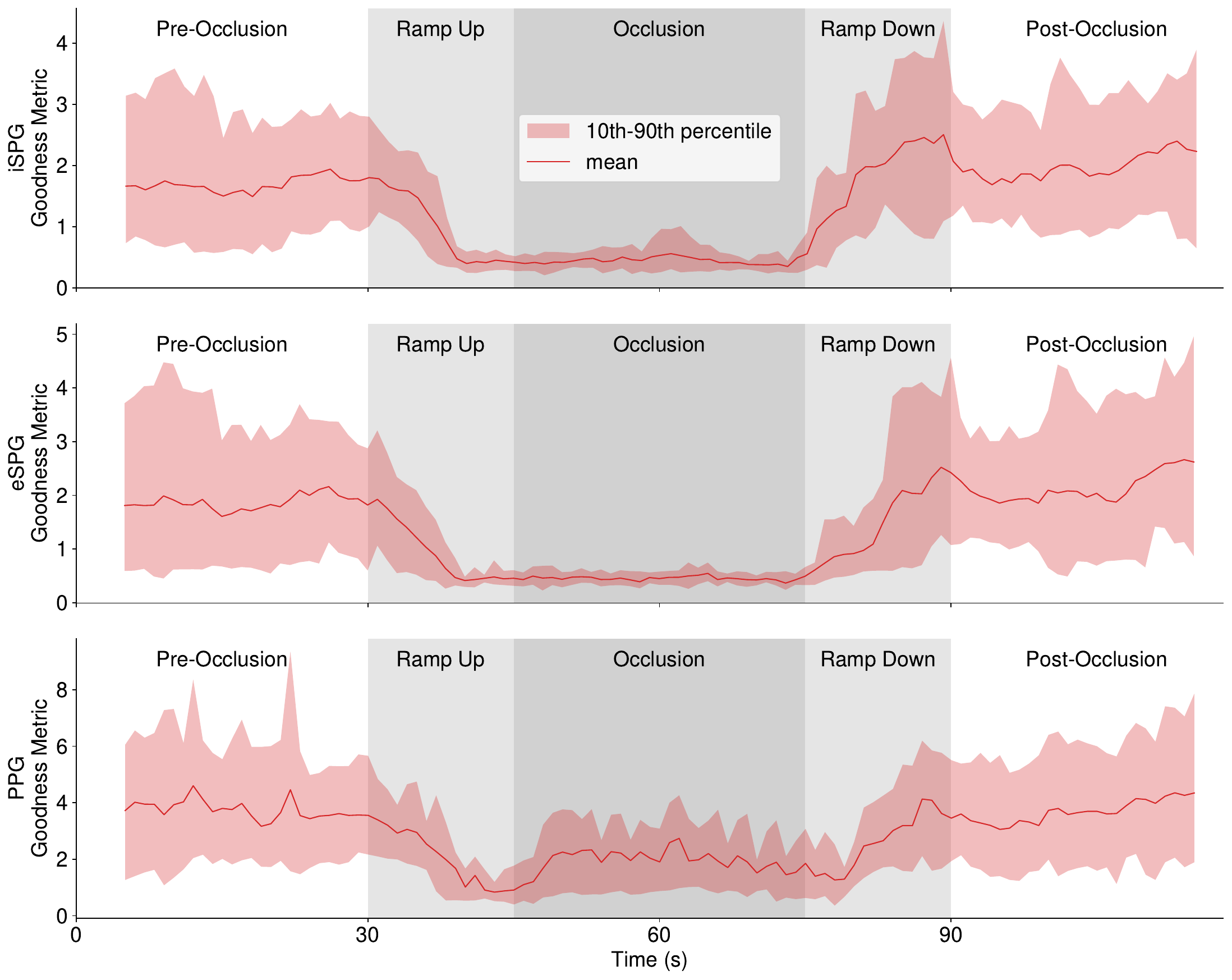}
    \caption{\textbf{eSPG detects cuff-induced vascular occlusion, thus demonstrating its ability to detect hemodynamic changes.} We plot the "Goodness Metric"~\cite{kumar2015distanceppg} of estimated eSPG, iSPG~\cite{olazabal2021beat}, and the GT PPG signals for 10 participants (short-range). Here, the solid line represents the mean across the participants, while the spread captures the variation within the $10^{th}-90^{th}$ percentile. The $2~min$ recording can be decomposed into 5 parts: the pre-occlusion still phase ($30~secs$), the pressure ramp-up phase ($15~secs$), the full-occlusion phase with steady $140~mm$~Hg cuff pressure ($30~secs$), the pressure ramp-down phase ($15~secs$), and the post-occlusion phase ($30~secs$). As the pressure in the cuff increases, the Goodness Metric drops rapidly, with the eSPG nearly flatlining under full occlusion. This trend is observed across the dataset, as evidenced by the narrow percentile spread.}
    \label{fig:cuff_occlusion}
\end{figure}

\par \Cref{fig:cuff_occlusion} illustrates the pan-dataset trend across the 5 phases. To derive this plot, we calculate the "Goodness Metric"~\cite{kumar2015distanceppg} over a $10~sec$ rolling window, with a stride of $1~sec$. Thus, by plotting the mean Goodness Metric and its spread ($10^{th}-90^{th}$ percentile), we infer that eSPG truly can image changes in blood flow. Furthermore, eSPG has the narrowest spread within the full occlusion region across the modalities. This is a strong indication of the method's sensitivity to dynamic speckle and its high invariance to static speckle, such as that arising from static surface scattering.

\newpage
\subsection{Heart Rate Variability (HRV) and HRV-based Respiratory Rate (RR)}
\label{subsec:hrv}

\par HRV (pulse rate variability, PRV) offers insight into the autonomic nervous system and its two main branches: the sympathetic nervous system, which drives fight-or-flight responses, and the parasympathetic nervous system, which governs rest-and-digest functions~\cite{electrophysiology1996heart,shaffer2017overview}. This makes HRV a promising biomarker for longitudinal studies into wellbeing~\cite{kim2018stress} and cognitive load~\cite{harish2025cogphys}. Accordingly, we extend our analysis beyond HR and explore heart rate variability as another axis of evaluation.

\begin{figure}[!ht]
    \centering
    \includegraphics[width=\textwidth]{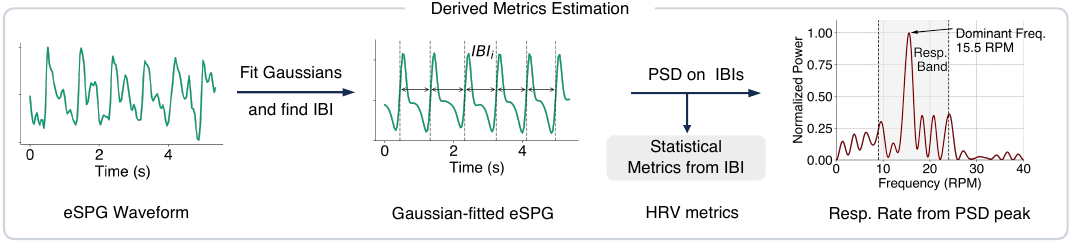}
    \caption{\textbf{Heart rate variability (HRV) and respiratory rate (RR) can be estimated from the statistical and spectral properties of the eSPG signal.} We denoise the estimated eSPG signals via an analysis-by-synthesis workflow. We fit a Gaussian mixture at each pulse and estimate the interbeat interval (IBI) from the synthesized waveform. The IBI signals can be used to calculate downstream HRV statistics. Further, its PSD also gives us the RR.}
    \label{fig:hrv_approach}
\end{figure}

\subsection*{Metrics}

\par We compute HRV from the interbeat interval (IBI; in $ms$). The standard mathematical approaches include calculating the standard deviation of normal-to-normal Intervals (SDNN) and the root mean square of successive differences (RMSSD). Formally, they are defined as:

\begin{align}
    \text{IBI}_i &= \text{peak}_{i} - \text{peak}_{i-1}, \\
    \text{SDNN} &= \sqrt{\frac{1}{N} \sum_{i=1}^{N} (\text{IBI}_i - \overline{\text{IBI}})^2}, \qquad \text{where}~\overline{\text{IBI}} = \frac{1}{N} \sum_{i=1}^{N} \text{IBI}_i, \\
    \text{RMSSD} &= \sqrt{\frac{1}{N-1} \sum_{i=1}^{N-1} (\text{IBI}_{i+1} - \text{IBI}_i)^2}
\end{align}

\par The HRV metrics are computed directly from the temporal waveforms and can be easily corrupted by motion, illumination changes, and large imaging distances. This necessitates high SNR. Commercial products achieve this by computing HRV during sleep, meditation, or other stationary periods~\cite{miller2022validation,whoop2026hrv,oura2025hrv,apple2026hr,garmin2026hrv}. We adopt a similar approach and limit HRV estimation to the \textbf{still} case in the short-distance experiments.

\subsection*{Gaussian Waveform Fitting and HRV Filtering}

\par The necessity of high SNR for HRV computation poses a challenge to remote signals such as eSPG and even iSPG. Simple operations, such as detrending and bandpass filtering, are insufficient. This leads to HRV metrics performing poorly, with SDNN and RMSSD explaining little of the observed variance.  Therefore, the estimated remote waveforms must be preprocessed to enable accurate peak detection and clean the interbeat sequence, $IBI_i$. 

\par To mitigate noise, we denoise the signals using an analysis-by-synthesis approach as illustrated in \cref{fig:hrv_approach}. We fit a parametric pulse model and use its reconstruction as the clean signal. To keep the optimization simple, we model each pulse of the remote plethysmography signal as a mixture of $K=4$ Gaussians, each with a learnable amplitude, center, and width, which together form our ``template''. This template is repeated at each pulse location, with each Gaussian scaled independently. To place the pulses, we include a learnable HR estimate initialized from the power spectral density (PSD) of the raw signal and use it to evenly space the pulse locations. A per-pulse jitter is then trained along with the HR estimate. We define the synthesis equation mathematically below, with the following trainable parameters:
\begin{enumerate}
    \item The amplitude ($A_k$), center ($\mu_k$), and width of each Gaussian ($\sigma_k$).
    \item A scaling factor for each Gaussian per pulse in the signal ($S_{i,k}$).
    \item The HR estimate ($h_{\text{learn}}$), initialized from the PSD of the raw signal.
    \item The jitter used to determine the pulse location ($\Delta\tau_i$). 
\end{enumerate}

\begin{align}
\tau_i &= i \cdot \frac{60}{h_{learn}} + \Delta\tau_i \\
z_t &= \sum_{i=1}^{P} \sum_{k=1}^{K} S_{i,k} A_k \exp\!\left(-\frac{(t - \mu_k - \tau_i)^2}{2\sigma_k^2}\right)
\end{align}

\par We train these models using two loss functions and four regularizers. The two loss functions ensure faithful reconstruction of the signal in the time and frequency domains, respectively. They consist of the mean squared error (MSE) loss applied between the detrended filtered raw signal ($y_t$) and the Gaussian-fitted reconstruction ($z_t$), and the mean absolute error (MAE) loss applied between the normalized PSDs ($\Phi_f$) in the passband, $F=[40~BPM-180~BPM]$. To improve signal fidelity, we introduce four regularizers, each designed to introduce an inductive bias into training. Among the four are three smoothness terms for the pulse template ($p_t$; with length $T_p$), the IBI sequence (for $P$ pulses), and the pulse location jitter. Finally, we include an HR anchor term to keep the HR computed from the pulse locations close to the HR computed from the PSD of the raw signal ($h_{raw}$). 

% Losses
\begin{align}
\mathcal{L}_{\text{time}} &= \frac{1}{T} \sum_{t} \left( z_t - y_t \right)^2 \\
\mathcal{L}_{\text{freq}} &= \frac{1}{|F|} \sum_{f \in F} \left| \Phi^{z}_f - \Phi^{y}_f \right|,
\qquad \text{where }\Phi^{y}_f = \frac{|\mathcal{F}\{y\}_f|^2}{\sum_{f' \in F} |\mathcal{F}\{y\}_{f'}|^2}
\end{align}

% Regularizers
\begin{align}
\mathcal{R}_{\text{tmpl}} &= \frac{1}{T_p} \sum_{t} \left| p_{t+1} - p_t \right|,
\qquad \text{where }p_t = \sum_{k=1}^{K} A_k \exp\!\left(-\frac{(t - \mu_k)^2}{2\sigma_k^2}\right) \\
\mathcal{R}_{\text{IBI}} &= \frac{1}{P} \sum_{i} \left| \text{IBI}_{i+1} - \text{IBI}_i \right|,
\qquad \text{where }\text{IBI}_i = \tau_{i+1} - \tau_i \\
\mathcal{R}_{\text{jit}} &= \frac{1}{P} \sum_{i} \left( \Delta\tau_{i+1} - \Delta\tau_i \right)^2 \\
\mathcal{R}_{\text{HR}} &= \left( \frac{60}{\overline{\text{IBI}}} - h_{raw} \right)^2
\end{align}

% Total
\begin{equation}
\mathcal{L} = \mathcal{L}_{\text{time}} + \lambda_{\text{freq}} \mathcal{L}_{\text{freq}}
+ \lambda_{\text{HR}} \mathcal{R}_{\text{HR}} + \lambda_{\text{IBI}} \mathcal{R}_{\text{IBI}}
+ \lambda_{\text{jit}} \mathcal{R}_{\text{jit}} + \lambda_{\text{tmpl}} \mathcal{R}_{\text{tmpl}}
\end{equation}

\par To calculate the HRV, we need to extract the individual beat locations for the denoised signal and the GT. First, we upsample the signals to $1000~Hz$ using cubic spline interpolation to improve temporal resolution. Then, we localize the beats by tracking fiducial points on the waveform - specifically, the steepest upstroke of the systolic peak (i.e., the maximum of the signal derivative). The PPG upstroke is more robust to noise and enables more precise localization than the systolic peak~\cite{mejia2022photoplethysmography}. We compute a smooth version of the derivative using a Savitzky-Golay filter~\cite{savitzky1964smoothing} with a $7$-sample window and a $ 4^{th}$-order polynomial. The peaks in the signal derivative are then used to extract the IBIs and derive SDNN and RMSSD. To mitigate the effects of noise in peak detection and imperfect signal fitting, we reject outliers in the IBI sequence by limiting values to within the interquartile range (IQR). Correspondingly, the rejected IBI values are masked out, with the preceding neighbor also masked when computing RMSSD.

\subsection*{Evaluation Scheme}

\par We employ a leave-one-participant-out (LOPO) evaluation scheme on the 10 recordings. For each participant, we choose the best hyperparameter by evaluating the performance of the various models on the remaining 9 participants. The best hyperparameter set is selected from 30 possible sets. Specifically, we vary $\lambda_{\text{freq}}$, $\lambda_{\text{HR}}$, $\lambda_{\text{IBI}}$, and $\lambda_{\text{jit}}$, while keeping $K=4$, $\lambda_{tmpl}=1$, $n_{iters}=2500$, and $lr=10^{-3}$. The learnable HR is initialized using the same HR estimation workflow from our main paper (Section 3).

\begin{table*}[ht]
\centering
\caption{\textbf{HRV error per trial}. We report the Standard Deviation of NN intervals (SDNN) and the Root Mean Square of Successive Differences (RMSSD) for our method (eSPG) and the baseline (iSPG). $\Delta$ is the error calculated as the estimate minus the ground truth. All metrics - SDNN, RMSSD, and their errors - are measured in milliseconds($ms$). Each participant's Gaussian-fit estimate uses the loss weights picked without them, by leave-one-participant-out on the MAE of RMSSD. Weights are $\lambda_{\text{freq}}$, $\lambda_{\text{HR}}$, $\lambda_{\text{IBI}}$, $\lambda_{\text{jit}}$.}
\begin{tabular}{l c c c c c c c c}
\toprule
 & \multicolumn{2}{c}{GT} & \multicolumn{3}{c}{\textbf{eSPG} (ours)} & \multicolumn{3}{c}{iSPG} \\
\cmidrule(lr){2-3} \cmidrule(lr){4-6} \cmidrule(lr){7-9}
Participant & SDNN & RMSSD & Weights & $\Delta$SDNN & $\Delta$RMSSD & Weights & $\Delta$SDNN & $\Delta$RMSSD \\
\midrule
p1 & $47.3$ & $43.7$ & $0, 0, 0, 10$ & $-9.3$ & $-8.4$ & $0, 10, 0, 10$ & $-3.7$ & $-0.6$ \\
p3 & $32.0$ & $43.2$ & $0, 0, 0, 10$ & $-9.0$ & $-13.7$ & $0, 10, 0, 10$ & $-2.8$ & $-7.8$ \\
p4 & $70.1$ & $41.7$ & $0, 0, 0, 10$ & $+8.0$ & $+40.2$ & $0, 10, 0, 10$ & $+2.7$ & $+13.0$ \\
p5 & $118.7$ & $139.0$ & $0, 0, 0, 10$ & $+7.5$ & $+17.1$ & $0, 10, 0, 10$ & $+1.9$ & $+6.0$ \\
p6 & $44.0$ & $58.6$ & $0, 0, 0, 10$ & $-7.8$ & $-9.1$ & $0, 10, 0, 10$ & $-2.4$ & $-4.6$ \\
p7 & $96.7$ & $97.7$ & $0, 0, 0, 10$ & $-18.1$ & $-27.2$ & $0, 10, 0, 10$ & $-15.0$ & $-19.6$ \\
p8 & $41.7$ & $36.2$ & $0, 0, 0, 10$ & $-4.1$ & $+1.6$ & $0, 10, 0, 10$ & $-4.1$ & $-0.1$ \\
p9 & $38.2$ & $38.2$ & $10, 10, 0, 10$ & $+16.7$ & $+32.1$ & $0, 10, 0, 10$ & $+29.6$ & $+33.9$ \\
p10 & $44.2$ & $54.2$ & $10, 0, 0, 10$ & $+64.8$ & $+114.2$ & $0, 10, 0, 0$ & $+28.7$ & $+52.9$ \\
p11 & $36.8$ & $23.7$ & $0, 0, 0, 10$ & $+3.3$ & $+6.6$ & $0, 10, 0, 10$ & $+0.6$ & $+14.0$ \\
\midrule
Mean Error &  &  &  & $+5.2$ & $+15.3$ &  & $+3.5$ & $+8.7$ \\
MAE &  &  &  & $14.9$ & $27.0$ &  & $9.1$ & $15.2$ \\
\bottomrule
\end{tabular}
\label{tab:hrv_still}
\end{table*}

\par From \cref{tab:hrv_still}, we observe that both eSPG and iSPG achieve sub-$10~ms$ errors on SDNN and RMSSD across the 10 participants. Among them, iSPG appears to have fewer outliers and lower errors. Further, we find that Gaussian fitting improves HRV estimation by $4-5~ms$ compared to the filtered detrended waveform (iSPG: $13.2~ms$ SDNN \& $19.4~ms$ RMSSD; eSPG: $19.7~ms$ SDNN \& $25.1~ms$ RMSSD). In addition to the LOPO testing, we investigate the validity of the Gaussian fitting when restricted to a single hyperparameter set. Based on the most recurring hyperparameter set, we choose $(\lambda_{\text{freq}}=0, \lambda_{\text{HR}}=0, \lambda_{\text{IBI}}=0, \lambda_{\text{jit}}=10)$ for eSPG, and $(\lambda_{\text{freq}}=0, \lambda_{\text{HR}}=10, \lambda_{\text{IBI}}=0, \lambda_{\text{jit}}=10)$ for iSPG. Through this configuration, HRV errors are reduced by nearly half across all metrics (eSPG: $9.8~ms$ SDNN \& $17.5~ms$ RMSSD; iSPG: $6.9~ms$ SDNN \& $10.4~ms$ RMSSD). We surmise that Gaussian waveform fitting is beneficial for denoising raw rSPG signals but requires careful tuning to achieve optimal results.

\subsection*{HRV-based Respiratory Rate Estimation}

\par Human respiration modulates the physiological waveforms associated with cardiac activity, such as PPG, SPG, ECG, and their remote variants~\cite{liu2020comparison}. This modulation manifests in three forms: baseline wander, amplitude modulation, and frequency modulation~\cite{charlton2017extraction,mejia2022photoplethysmography}. The latter of the three arises from the respiratory sinus arrhythmia (RSA). While contact-based sensing permits respiratory rate (RR) estimation from any of the three sources~\cite{karlen2013multiparameter,birrenkott2017robust,hoogeveen2025three}, remote sensing is highly susceptible to amplitude and baseline corruption due to illumination changes and motion. That leaves us with frequency modulation as the primary source for RR estimation. 

\par Due to RSA-induced frequency modulation, the IBI decreases during inhalation and increases during exhalation. We can estimate the frequency of these rhythmic variations (i.e., the RR) by analyzing the PSD of the IBI sequence as shown in~\cref{fig:hrv_approach}. We compute the IBI sequence using the Gaussian fitting approach with the same set of hyperparameters. The chosen hyperparameters were optimized for IBI and HRV estimation. Hence, we surmise that they would retain the RSA-induced variations. To mitigate low-frequency effects from non-respiratory variations, we apply a first-order derivative filter~\cite{hoogeveen2025three} on the IBI sequence. Finally, we employ IQR filtering to filter out irregular IBI jumps accentuated by the derivative filter. The PSD is then calculated on this filtered sequence using the Lomb-Scargle periodogram, as it is best suited for irregularly sampled weak periodic signals~\cite{lomb1976least,scargle1982studies,zechmeister2009generalised,vanderplas2018understanding}.

\begin{table}[ht]
\centering
\caption{\textbf{Respiratory rate (RR) modulates the heartbeat and can be extracted from plethysmography signals.} We choose the best hyperparameter fit using the LOPO strategy from our HRV analysis. The RR estimated from the pulse oximeter is used as our GT measure. All values are measured in respirations per minute ($RPM$). The method columns are $\Delta$, estimate minus ground truth.}
\begin{tabular}{l c c c}
\toprule
Trial & GT & \textbf{eSPG} (ours) & iSPG \\
\midrule
p1 & $12.6$ & $+2.4$ & $+0.0$ \\
p3 & $17.8$ & $-1.9$ & $+0.3$ \\
p4 & $20.3$ & $+1.6$ & $+1.5$ \\
p5 & $17.2$ & $-1.6$ & $-1.4$ \\
p6 & $18.5$ & $+0.0$ & $+0.0$ \\
p7 & $8.6$ & $+1.0$ & $+0.9$ \\
p8 & $15.0$ & $+0.0$ & $+0.0$ \\
p9 & $18.0$ & $-3.0$ & $+0.5$ \\
p10 & $18.1$ & $+0.1$ & $+0.0$ \\
p11 & $8.5$ & $+0.0$ & $+0.0$ \\
\midrule
Mean &  & $-0.1$ & $+0.2$ \\
MAE &  & $1.2$ & $0.5$ \\
\bottomrule
\end{tabular}
\label{tab:resp_from_hrv}
\end{table}

\par From \cref{tab:resp_from_hrv}, we observe that RR, measured as respirations per minute ($RPM$), can be estimated with under 2 $RPM$ error for a majority of the samples. Although better for HRV, iSPG is worse than eSPG for RR estimation. This suggests that while eSPG is worse at extracting the exact IBI values, it is better at capturing the trend, i.e., the frequency modulation caused by the respiration. When compared to the RR estimated from the raw waveforms (eSPG: $1.7~RPM$; iSPG: $1.9~RPM$), Gaussian fitting improves RR estimation, with iSPG improving by $\approx4\times$.

\subsection*{Limitation and Improvements}

\par As seen from \cref{tab:hrv_still,tab:resp_from_hrv} and the accompanying analysis, the choice of hyperparameters is key to model performance. In our study, we have shown the potential to extract more complex physiological markers, such as HRV and RR, from remote speckle-based sensing, such as eSPG and iSPG. However, our dataset is limited to 10 samples, so the best-performing hyperparameters may not generalize to a broader population. Hence, we advocate for a large-scale future study, preferably with ECG as a reference, to quantify the statistical validity and clinical applicability of using rSPG signals with Gaussian waveform fitting.

\newpage
\subsection{Bland-Altman Plots}

\par We use Bland-Altman plots to compare eSPG and iSPG, and further analyze their agreement with the ground-truth fingertip PPG recordings. We isolate the effects of distance and imaging conditions by separating mid- and long-range experiments from short-range experiments. The latter experiment set tests the system's robustness to various external interfering sources, such as motion, ambient light, and sensitivity to the choice of laser illumination. 

\par
\begin{figure}[!htbp]
    \centering
    \includegraphics[width=\linewidth]{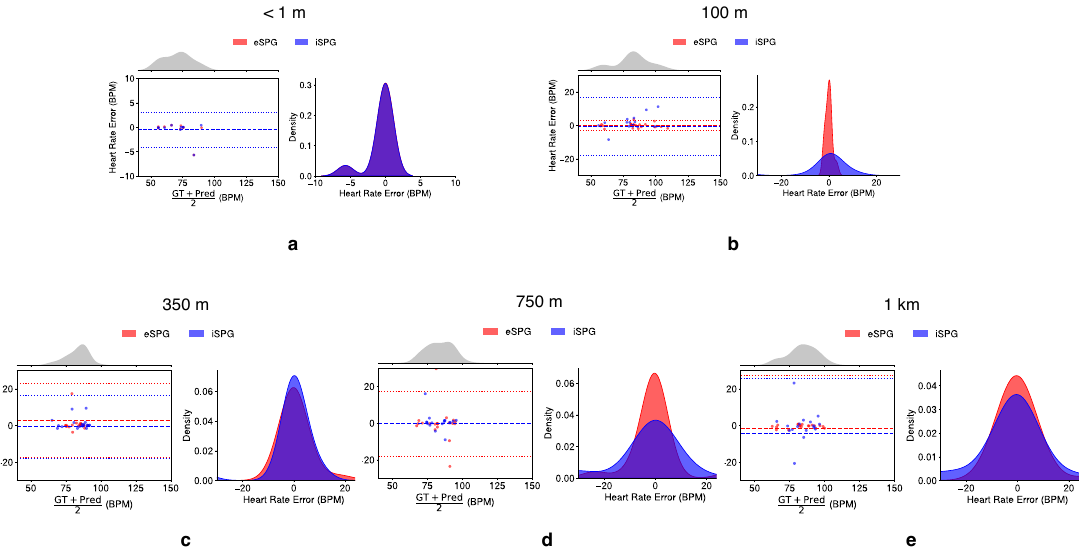}
    \caption{\textbf{eSPG is more consistent across different imaging distances, each captured using a different focusing optic.} \textbf{a,} For short distances ($<1~m$) and a $12.5~mm$ lens, the error distributions of eSPG and iSPG overlap almost perfectly, showing that the two methods both perform equally well.
    \textbf{b,} At $100~m$ with a $135~mm$ lens, eSPG performs significantly better than iSPG. iSPG achieves lower error on some samples, but eSPG is more consistent, with lower error across the full set. \textbf{c, d, e,} In long-range sensing, with a telescope configuration and laser placed at $60~m$, eSPG and iSPG exhibit competitive performance. The eSPG estimates increasingly outperform iSPG at greater imaging distances.}
    \label{fig:ba_distance}
\end{figure}

\begin{figure}[!htbp]
    \centering
    \includegraphics[width=\linewidth]{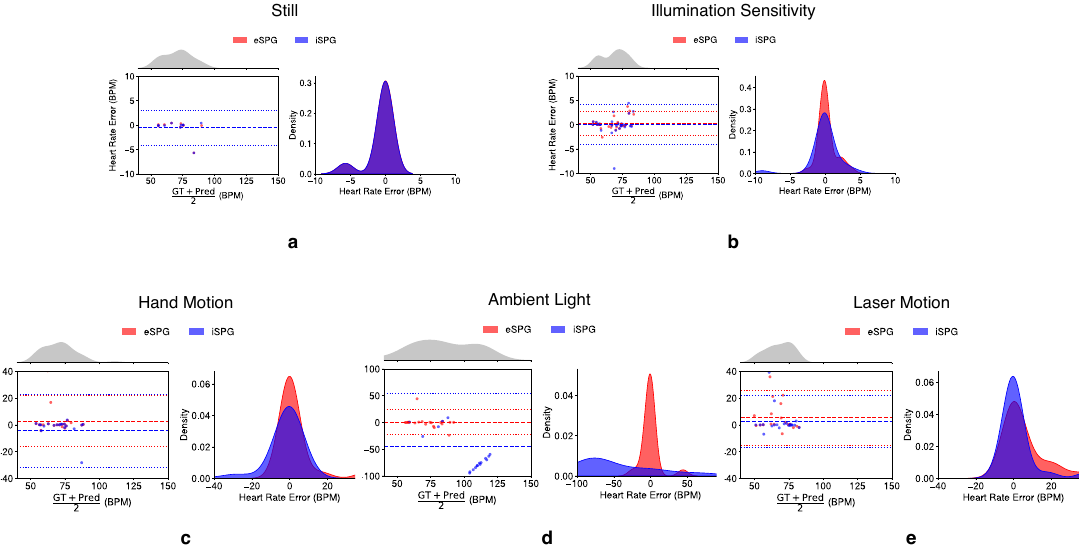}
    \caption{\textbf{eSPG-based HR estimation surpasses prior work across different imaging conditions.} All samples are recordings of the palm at short imaging distances ($<1~m$) with a $12.5~mm$ lens. \textbf{a,} When the palm remains still, the speckle fluctuations can be sensed equally well with event- and frame-based cameras. \textbf{b,} eSPG is more resilient to changes in laser illumination. This includes changes to the beam profile (Gaussian or uniform) and to the laser power (sub-$mW$ laser pointer). \textbf{c,} Event-based sensing is more robust to motion-induced corruption when compared to frame-based methods. The error distributions are visibly more compact, with fewer individual outliers. \textbf{d,} Imaging in extreme ambient light favors event cameras due to their inherently high dynamic range and relative insensitivity to bright static illumination. \textbf{e,} Moving the laser corrupts the speckle fluctuations captured by both cameras. In this case, iSPG performs marginally better, though both methods yield similar high-error outliers.}
    \label{fig:ba_short_all}
\end{figure}

 \Cref{fig:ba_distance} demonstrates that eSPG remains a stable sensing method to estimate heart rate across a wide range of imaging distances, imaging optics, and GT HRs. The eSPG errors are consistently more compact, as illustrated by $\cdots$ lines (denoting $\pm1.96\times\sigma$). Specifically, we note that the error histogram for eSPG is $>6\times$ narrower than that of iSPG for the $100~m$ imaging setup.
The superior performance of eSPG is further evidenced by \cref{fig:ba_short_all}. We observe that eSPG is robust to external sources of interference such as changes in laser illumination, hand motion, and extreme ambient light. 

\newpage
\section{Participant Demographics}

\begin{figure}[ht]
    \centering
    \includegraphics[width=0.7\linewidth]{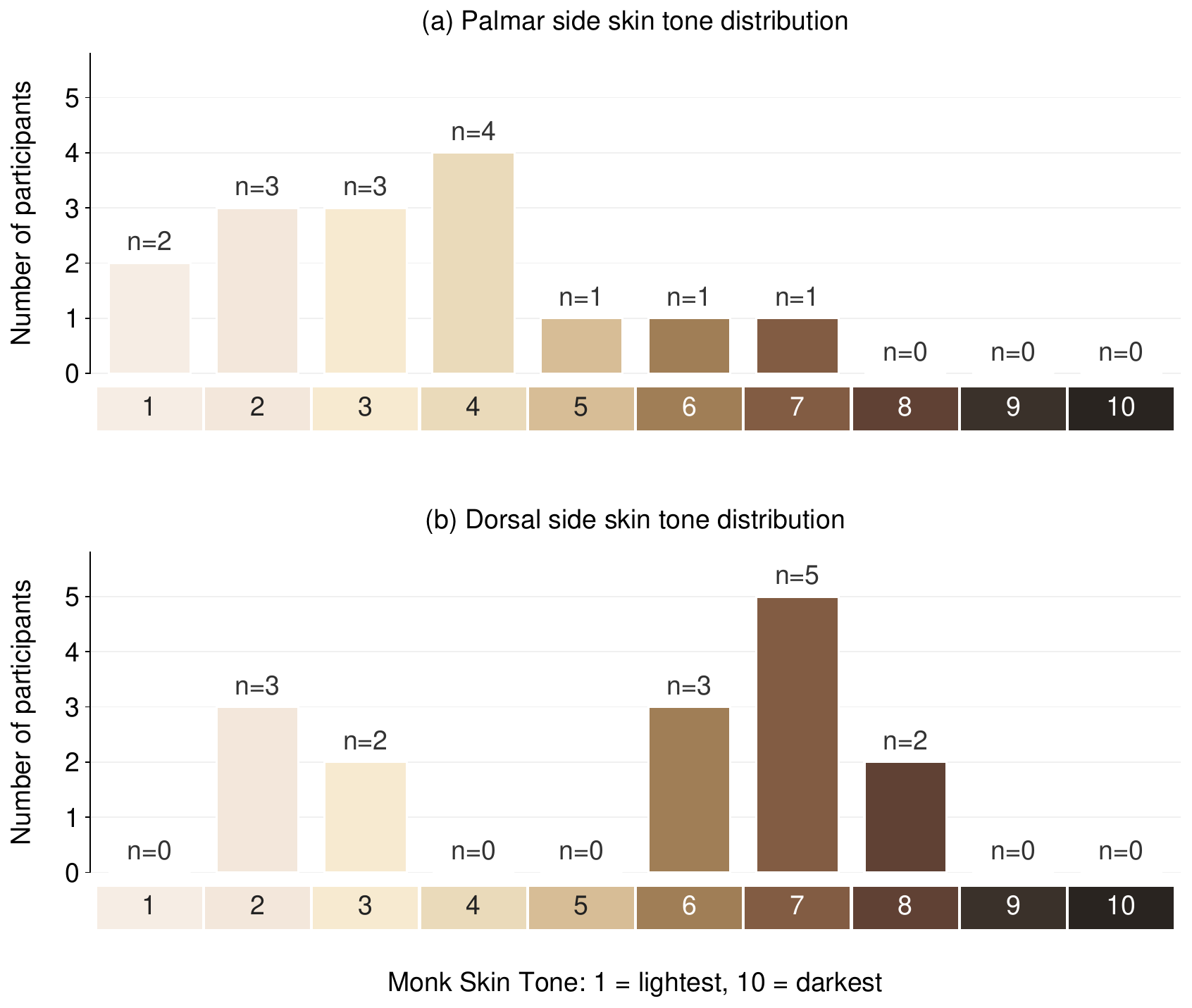}
    \caption{Participant distribution across Monk Skin Tone Scale \cite{monk2019}. The palmar side skin tone is measured at the palm-wrist intersection, while the dorsal side is measured on the back side of the palm.}
    \label{fig:monk_scale}
\end{figure}

We recorded data from a total of 15 participants across a range of imaging conditions. Our dataset includes participants aged 20-35, with 9 male and 6 female participants. \Cref{fig:monk_scale} shows the distribution of participants across skin tones measured by the Monk Skin Tone Scale~\cite{monk2019}.

% \newpage
% \bibliography{references}

\newpage
\bibliography{references}

\end{document}